\documentclass[10pt,twocolumn,letterpaper]{article}

\usepackage{cvpr}              %

\usepackage{graphicx}
\usepackage{amsmath}
\usepackage{amssymb}
\usepackage{booktabs}
\usepackage{tabularx}
\usepackage{axessibility}

\usepackage{tabularx} 
\newcolumntype{L}{>{\raggedright\arraybackslash}X}
\newcolumntype{R}{>{\raggedleft\arraybackslash}X}
\newcolumntype{C}{>{\centering\arraybackslash}X} 

\usepackage{color, colortbl} %
\definecolor{LightGray}{rgb}{0.9,0.9,0.9}

\usepackage{multirow}

\usepackage{graphicx} 
\usepackage{animate}

\usepackage{pifont}

\usepackage{graphbox}

\usepackage{enumitem}
\setitemize{noitemsep,topsep=0pt,parsep=0pt,partopsep=0pt}

\usepackage[pagebackref,breaklinks,colorlinks]{hyperref}
\usepackage{color}
\usepackage{xcolor}
\newcommand{\rev}[1]{{\color{black} #1}}

\usepackage[capitalize]{cleveref}
\crefname{section}{Sec.}{Secs.}
\Crefname{section}{Section}{Sections}
\Crefname{table}{Table}{Tables}
\crefname{table}{Tab.}{Tabs.}

\def\cvprPaperID{1762} %
\def\confName{CVPR}
\def\confYear{2022}

\definecolor{somegray}{rgb}{0.5, 0.5, 0.5}
\newcommand{\darkgrayed}[1]{\textcolor{somegray}{#1}}
\makeatletter
\newcommand*\titleheader[1]{\gdef\@titleheader{#1}}
\AtBeginDocument{%
  \let\st@red@title\@title
  \def\@title{%
    \vskip-3em
    \bgroup\normalfont\large\centering\@titleheader\par\egroup
    \vskip1.5em\st@red@title}
}

\makeatother

\titleheader{\darkgrayed{This paper has been accepted for publication at the\\
IEEE Conference on Computer Vision and Pattern Recognition (CVPR), New Orleans, 2022.
\copyright IEEE}}

\makeatletter
\let\@oldmaketitle\@maketitle%
\renewcommand{\@maketitle}{\@oldmaketitle%
\centering
\begin{tabular}{ccc}
\includegraphics[width=0.325\textwidth]{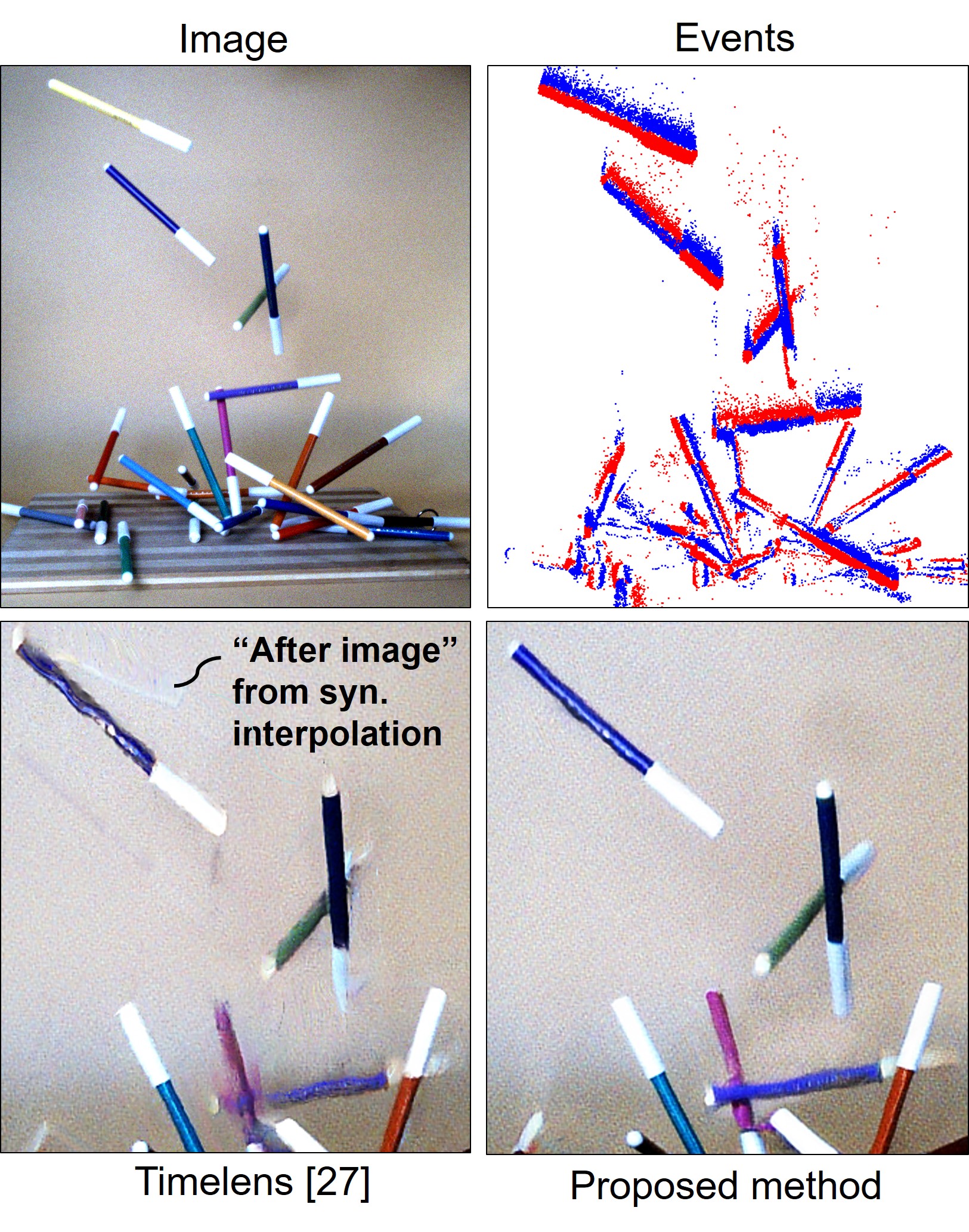}&
\includegraphics[width=0.318\textwidth]{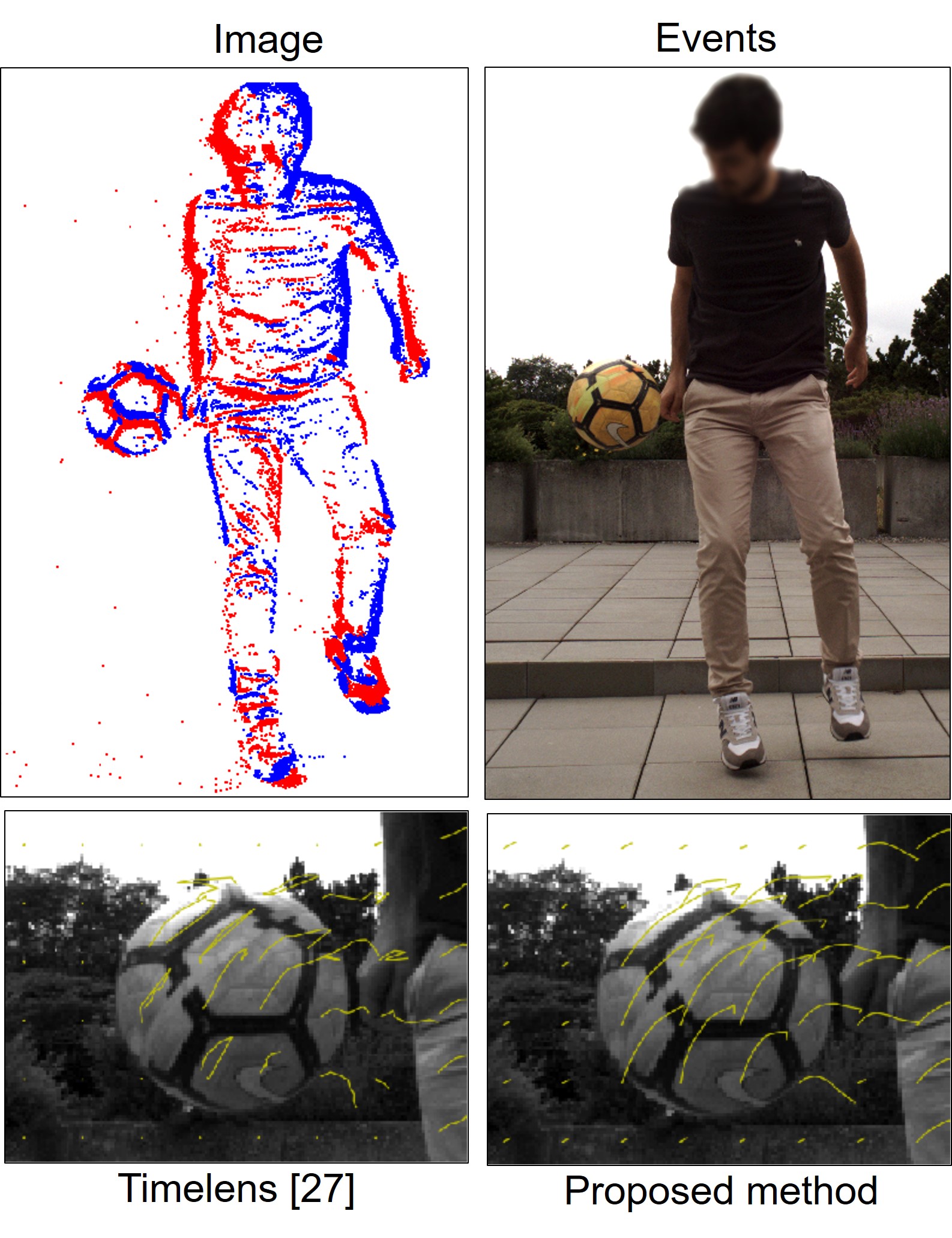}&
\includegraphics[width=0.307\textwidth]{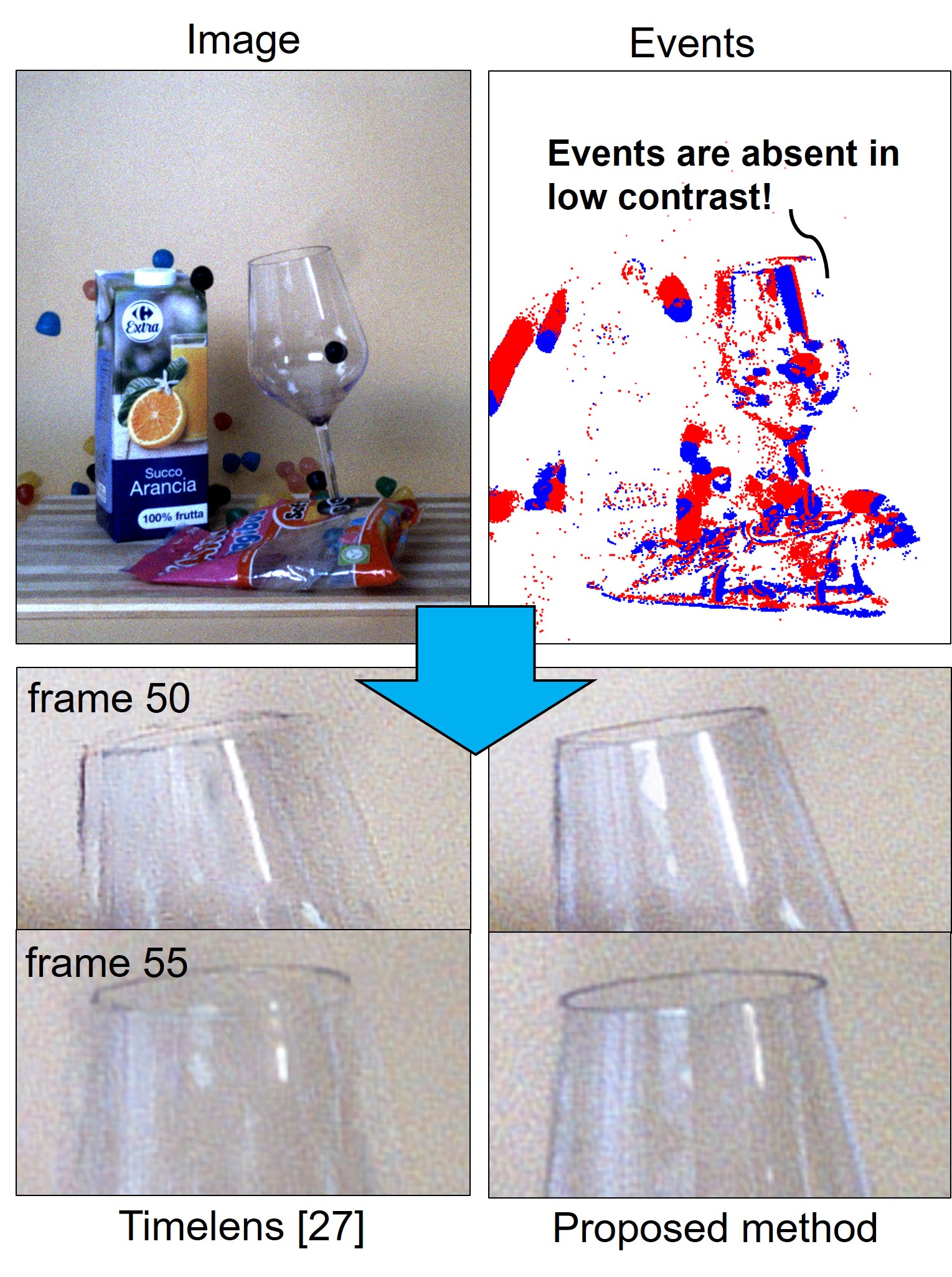}\\
(a)~robust fusion & (b)~fast \& temporally consistent motion & (c)~robustness to event sparsity
\end{tabular}
\captionof{figure}{%
Comparison to state-of-the-art event- and image-based video interpolation method Time Lens~\cite{tulyakov_gehrig_2021cvpr}. Our method makes a series of key innovations to address the limitations of current approaches. First, it uses feature-level multi-scale fusion which is robust to artifacts in the fused images (a). Second, it computes continuous flow, parametrized by splines, which have inherent temporal consistency (b, bottom right vs. left) and can be efficiently sampled, thereby significantly reducing computation for multi-frame interpolation (b). Finally, it combines images and events to generate flow, even where few events are triggered, thereby mitigating artifacts as in (c). 
\vspace{0.3cm}
}\label{fig:eyecatcher}
}%
\makeatother

\title{\rev{Time Lens++:} Event-based Frame Interpolation with \\Parametric Non-linear Flow and Multi-scale Fusion}

\begin{document}

\author{Stepan Tulyakov$^1$ \quad\quad Alfredo Bochicchio$^1$ \quad\quad Daniel Gehrig$^2$ \quad\quad Stamatios Georgoulis$^1$ \\ Yuanyou Li$^1$\quad\quad Davide Scaramuzza$^2$ \\

$^1$ Huawei Technologies, Zurich Research Center\\
$^2$ Dept. of Informatics, Univ. of Zurich and Dept. of Neuroinformatics, Univ. of Zurich and ETH Zurich\vspace{-3ex}
}
\maketitle
\begin{abstract}
\vspace{-2ex}
Recently, video frame interpolation using a combination of frame- and event-based cameras has surpassed traditional image-based methods both in terms of performance and memory efficiency. However, current methods still suffer from (i)~brittle image-level fusion of complementary interpolation results, that fails in the presence of artifacts in the fused image, 
(ii)~potentially temporally inconsistent and inefficient motion estimation procedures, that run for every inserted frame %
and (iii)~low contrast regions that do not trigger events, and thus cause events-only motion estimation to generate artifacts. %
Moreover, previous methods were only tested on datasets consisting of planar and far-away scenes, which do not capture the full complexity of the real world. In this work, we address the above problems by introducing multi-scale feature-level fusion and computing one-shot non-linear inter-frame motion---which can be efficiently sampled for image warping---from events and images. We also collect the first large-scale events and frames dataset consisting of more than 100 challenging scenes with depth variations, captured with a new experimental setup based on a beamsplitter.  %
We show that our method improves the reconstruction quality by up to 0.2 dB in terms of PSNR and up to 15\% in LPIPS score. 

\section*{Multimedia Material}
For videos, datasets and more visit \url{https://uzh-rpg.github.io/timelens-pp/}.

\end{abstract}

\section{Introduction}

High-speed videography has captured the imagination of the public by producing stunning slow-motion footage of high-speed phenomena~\cite{tulyakov_gehrig_2021cvpr,bao_cvpr2019, kalluri_arxiv2021}. %
While historically this was only possible with specialized and expensive equipment, today this technology is coming to our smartphones thanks to high-speed cameras and \textit{video frame-interpolation~(VFI)} techniques. 

VFI techniques generate high frame rate video by inserting intermediate frames between consecutive frames of a low framerate input video.
To this end, they estimate image changes in the blind time between consecutive frames, a task that remains challenging, especially in the presence of large displacements and non-linear motion. %
Most existing VFI methods rely on the information contained in the original video to estimate these changes. However, at low frame rates, image-based motion estimation does not accurately capture inter-frame motion, especially in presence of large and non-linear motion. %
While using high frame-rate cameras can alleviate this problem, they are typically expensive and produce excessive amounts of data, which cannot be recorded for an extended amount of time. For example, the Huawei~P40 can record only 0.5s of 720p video with 1920 fps, which fills up its 2GB frame buffer. %

\textit{Event cameras} address both of these problems. Instead of measuring synchronous frames of absolute brightness like standard cameras, they only measure asynchronous brightness changes at each pixel, which results in a sparse and asynchronous stream of \textit{events}, each encoding the position, time, and polarity (sign) of the measured change. This stream of events is simultaneously sparse, provides a low data bandwidth, and has a high temporal resolution on the order of microseconds capable of capturing high-speed phenomena, such as gunshots and popping water balloons\footnote{\url{https://youtu.be/eomALySSGVU}}.

The recently introduced Time Lens\cite{tulyakov_gehrig_2021cvpr} leverages this compressed stream of visual information by combining a traditional camera with an event camera to perform video frame interpolation. Its key innovation lies in combining the benefits of \textit{warping-based} and \textit{synthesis-based} interpolation approaches through an attention mechanism. Warping-based interpolation produces intermediate frames by warping frames of the original video using nonlinear motion estimated from events, while synthesis-based interpolation produces intermediate frames by ``adding'' intensity changes captured by inter-frame events to the frames of the original video. %
Time Lens combines these two approaches because they are complementary: while warping-based interpolation usually produces high-quality results, it suffers where motion estimation is unreliable due to violation of brightness constancy assumption. %
By contrast, the synthesis-based interpolation does not rely on brightness constancy and can easily handle objects with illumination changes such as, for example, fire, and water, however, it distorts fine texture due to the sparsity of events. %

\textbf{Motivation}. Despite its impressive performance compared with pure image-based methods,\footnote{\url{https://youtu.be/dVLyia-ezvo}} previous work suffers from several drawbacks.%
First, to combine warping- and synthesis-based interpolation it relies on image-level fusion which can fail in the presence of artifact in one of the inputs, as shown in Figure~\ref{fig:eyecatcher}(a), where ``after-image'' artifacts from synthesis interpolation are propagated to the final results. %
Second, it relies on non-parametric motion estimation, %
that runs independently for each inserted frame with a computational cost of $ O(N) $, where $ N $ is the number of inserted frames and produces potentially temporally inconsistent motion estimates. %
Third, to leverage information about non-linear motion it relies on events-only motion estimation, which leads to artifacts in low contrast areas without events (see Figure~\ref{fig:eyecatcher}(c)). %

This work addresses all of these open challenges. 
Our method makes a key innovation in terms of motion estimation, visualized in Fig. \ref{fig:motion_types}. Instead of using linear (b) or chunked linear flow from events (c), we use both images and intermediate events (a) to predict a continuous flow field (d). 
By doing so our flow method is inherently temporally consistent and can be efficiently reused for multi-frame insertion. 
Additionally, we introduce a novel multi-scale fusion module that fuses event and image features on feature-level instead of image-level, thereby limiting ghosting artifacts.

We make the following contributions in this work:
\begin{enumerate}[itemsep=0.25pt,topsep=0.25pt,parsep=0pt,partopsep=0pt]
    \item We introduce a novel \textit{motion spline estimator}, which produces non-linear continuous flow from events and frames. It is temporally consistent and can be efficiently sampled, enabling the interpolation of $ N $ intermediate frames with $ O(1) $ instead of $ O(N) $ computation. Moreover, leveraging images also produces accurate flow in the absence of events. %
    
    \item We introduce a \textit{multi-scale feature fusion} module with multiple encoders and joint decoder with a \textit{gated compression} mechanism that selects the most informative features from each encoder at each scale and improves fusion of warping- and synthesis-based interpolation results.

    \item We compare our approach on an existing dataset and a new \textit{large scale hybrid dataset} containing 123 videos collected with a beamsplitter setup that has temporally synchronized and aligned events and frames. We compare our method on multiple benchmarks including this new dataset and found an up to 0.2 dB improvement in terms of PSNR and up to 15\% improvement in perceptual score~\cite{zhang_2018} over the prior art, across datasets.
\end{enumerate}
\section{Related Work}
\label{sec:related_work}
\textbf{Frame-based} video interpolation is a well studied topic with an abundance of prior work~\cite{jiang_2018, niklaus_2017a, niklaus_2017b, niklaus_2018, niklaus_2020, paliwal_2020,park_2020,xue_2019}. 
It aims at reconstructing intermediate latent frames at arbitrary or fixed timestamps using consecutive frames of the original video, called keyframes.
Most image-based frame interpolation methods adopt one of four approaches:  While \emph{direct} approaches~\cite{kalluri_arxiv2021}, regress intermediate frames directly from keyframes, \emph{kernel-based} approach~\cite{niklaus_2017a, niklaus_2017b}, apply convolutional kernels to keyframes to produce the latent frame, and \emph{phase-based}\cite{meyer_2018} approaches, estimate the phase decomposition of the latent frame. The most popular approach is \emph{warping-based}\cite{jiang_2018, niklaus_2018, niklaus_2020, paliwal_2020,park_2020,xue_2019} and it explicitly estimates the motion between keyframes and then warps and fuses keyframes to produce the latent frame. This fusion is usually done on the image level, using visibility masks\cite{jiang_2018} or, more recently on the feature-level~\cite{bao_cvpr2019, niklaus_2020}. %
The majority of works tackling VFI are image-based and thus suffer from two major limitations. First, they rely on image-based motion estimation, which is only well defined when brightness constancy is satisfied. Secondly, they can not capture precise motion dynamics in the blind time between keyframes and often resort to a simplistic linear motion assumption. 

\textbf{Motion Estimation:} Indeed, motion estimation, has predominantly been studied in the linear case, where correspondences between pixels are assumed to follow linear trajectories. However, in the case of rotational camera ego-motion and non-rigid object motion, this assumption is usually violated. Only few works use more complex motion models, such as quadratic~\cite{xu_2019} or cubic~\cite{Chi20all,Chun12JNM}. While~\cite{xu_2019,Chi20all} directly regress polynomial coefficients, \cite{Chun12JNM} parametrize the pixel trajectories in terms of 2-D knots, using B-Splines. Fitting these non-linear models requires multiple frames and long time windows, and thus they still fail to model high-speed and non-linear motion between keyframes.

\textbf{Use of Additional Sensors:} To capture this motion, information from additional sensors with high temporal resolution can be used. In particular~\cite{paliwal_2020, gupta_2009} uses an auxiliary low resolution, high-speed camera to provide these additional cues, and combine them with high-resolution images. 
However, additional high-frame-rate image sensors increase data rate requirements. This is a fundamental limitation of frame cameras to capture high-speed motions, since they oversample the image, leading to wasteful data acquisition. 

\textbf{Event Cameras:} are sensors that ideally address this limitation since they mitigate this oversampling by only providing data at locations with intensity changes, and do this for each pixel independently. The works in \cite{lin_2020, yu_iccv2021, Han21iccv, Wang21iccv, tulyakov_gehrig_2021cvpr} have used an auxiliary event camera for VFI, demonstrating high accuracy and low bandwidth.  Time Lens\cite{tulyakov_gehrig_2021cvpr}, generates frame interpolations from a warping-based and synthesis-based module and fuses them on the image-level using learned alpha blending parameters. By combining the advantages of both, it can handle both regions with brightness constancy and those with illumination changes where optical flow is ill-defined. \cite{yu_iccv2021} improves the fusion part by performing progressive multi-scale, feature-level fusion. 

However,  \cite{yu_iccv2021} aligns keyframes to the latent frame using flow, computed from original keyframes, and thus can not capture non-linear inter-frame dynamic. Instead, \cite{tulyakov_gehrig_2021cvpr} computes flow from events, capturing non-linear inter-frame dynamic, and directly predicts a series of non-parametric linear flow between keyframes and the latent frame. However, this model does not take into account the continuous nature of events, and since flow is non-parametric it cannot be reused and must be recomputed. While this leads to a significant run-time increase, it also leads to temporal inconsistency, which manifests in wobbling textures. 
Finally, since flow is computed from events, it is sparse and inaccurate in low contrast regions that do not trigger events.

\begin{figure*}
\centering
\begin{tabular}{cccc}
     \includegraphics[height=3cm]{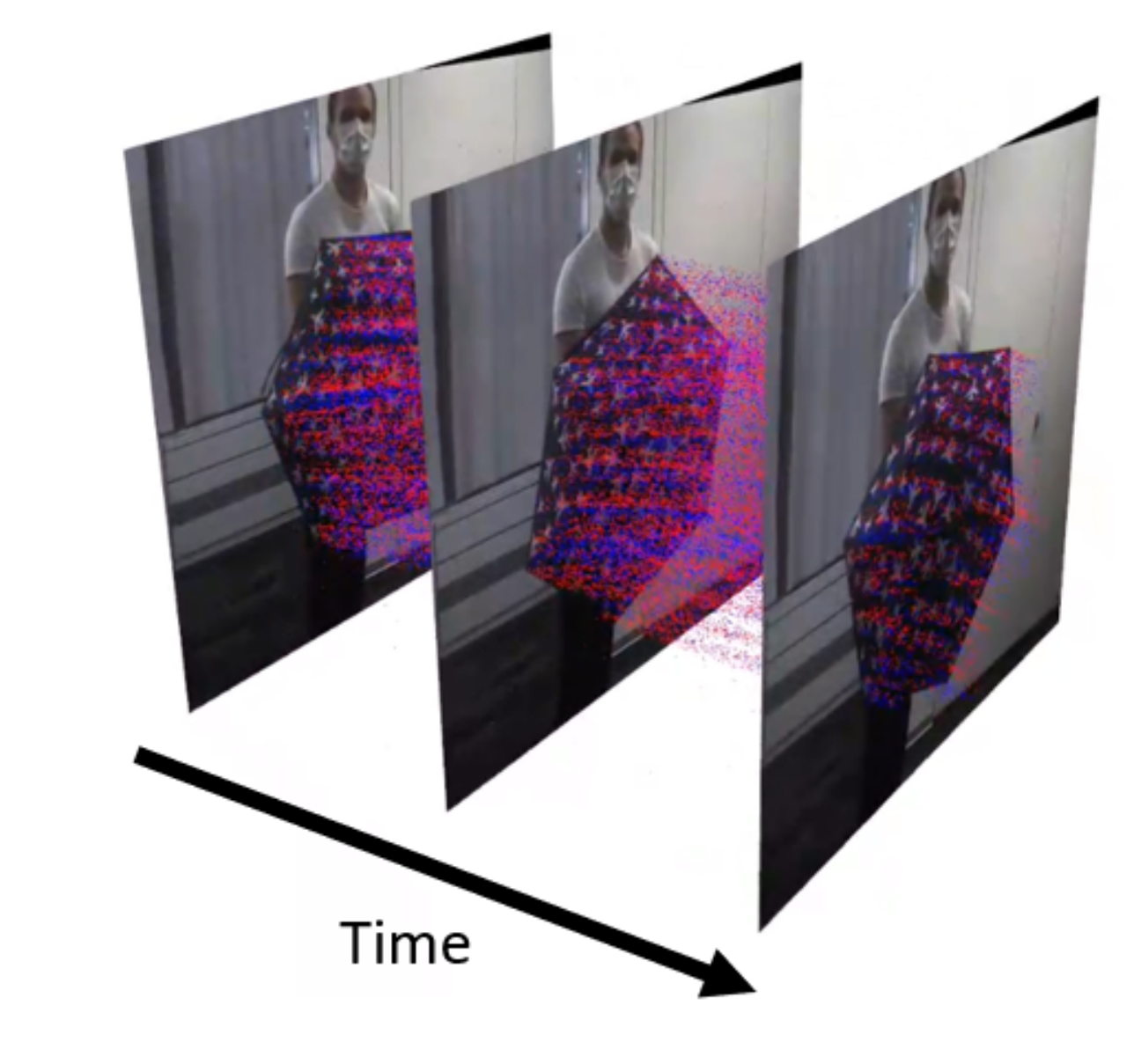}&
     \includegraphics[height=3cm]{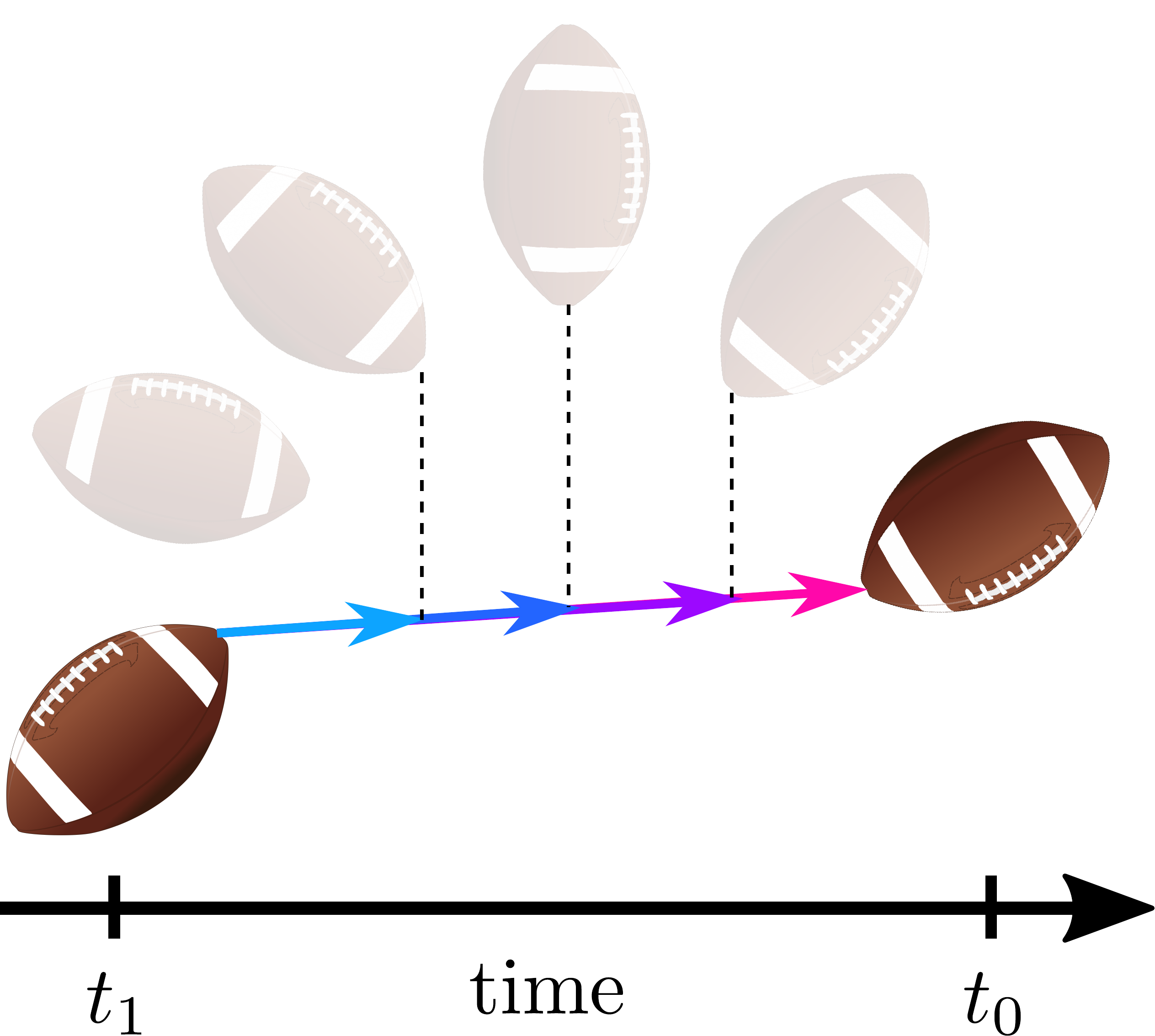}&
     \includegraphics[height=3cm]{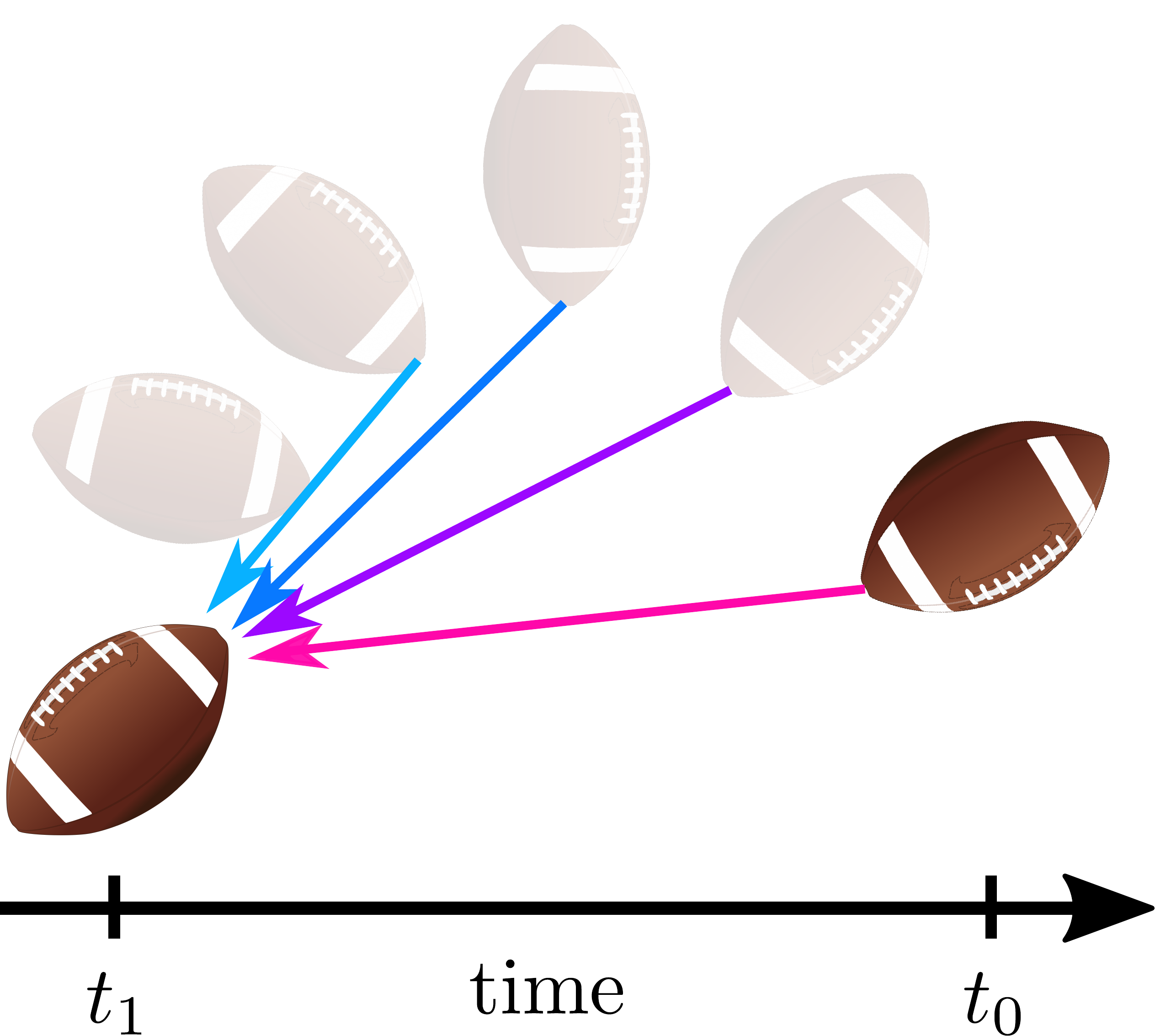}&
     \includegraphics[height=3cm]{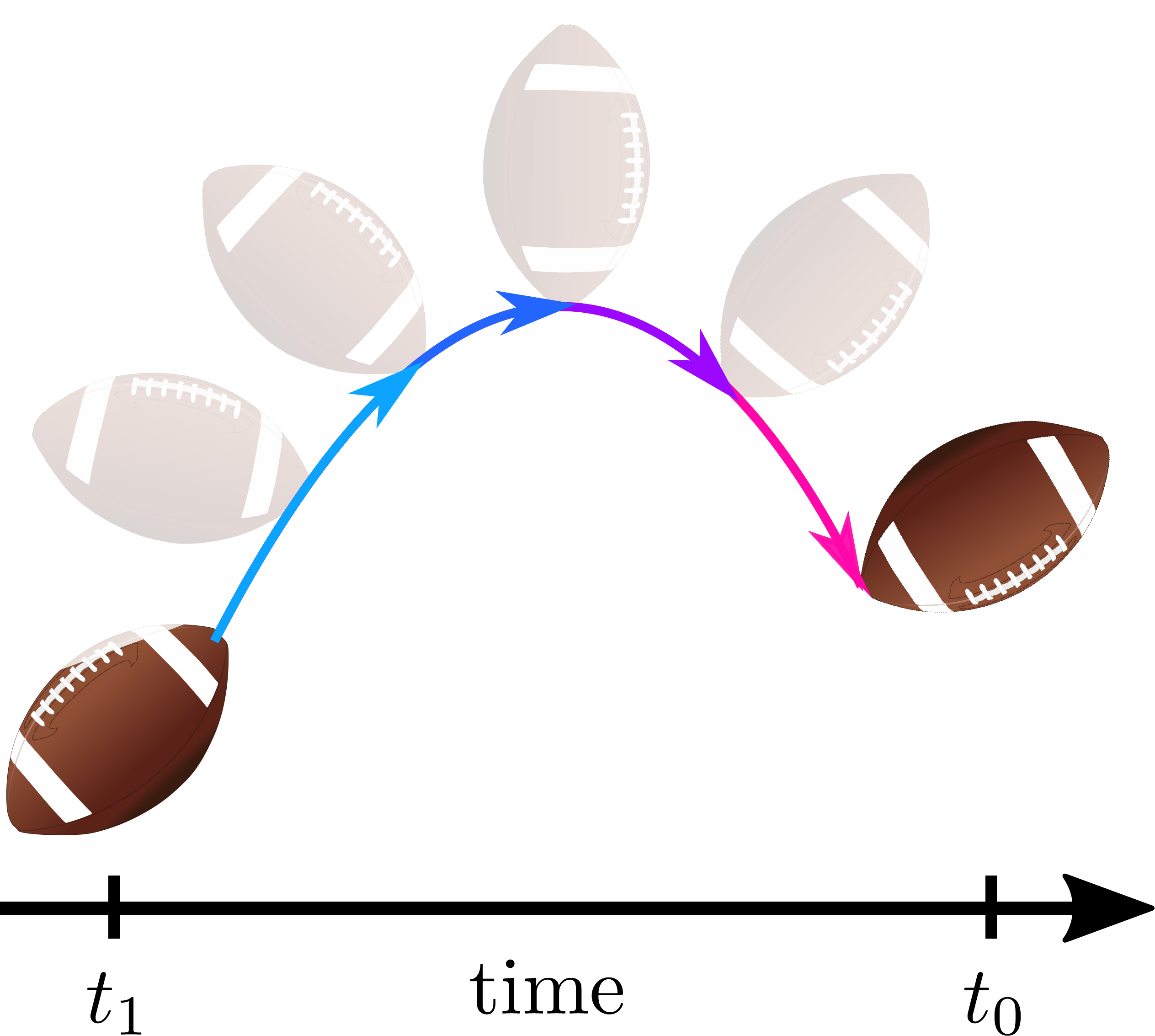}\\
     (a) events and frames & (b) linear motion &(c) non-parametric motion & (d) spline motion
\end{tabular}\caption{Different motion estimation types for video-frame interpolation~(VFI). More details are in Section \ref{sec:related_work}, ``motion estimation".} %
\label{fig:motion_types}
\end{figure*}

In this work, we combine the advantages of \cite{tulyakov_gehrig_2021cvpr} and \cite{yu_iccv2021} while addressing their limitations. Firstly, we make a key innovation in terms of motion estimation, visualized in Fig. \ref{fig:motion_types}. Instead of using linear (b) or chunked linear flow from events (c), we use both images and intermediate events (a) to predict a continuous flow field (d), based on cubic splines, similar to \cite{Chun12JNM}. See Fig.~\ref{fig:motion_types} for a visualization of their differences. 
The resulting flow has several advantages: \emph{(i)} it is non-linear, capturing high-speed and highly non-linear dynamics with microsecond resolution, \emph{(ii)} it is dense, thus producing flow, even where few events are present and \emph{(iii)} it can be efficiently reused during multi-frame insertion, resulting in low inference time and high temporal consistency.

\section{Method}
\label{sec:method}

\textbf{Problem formulation.} We are given as input proceeding $I_0$ and following $I_1$  key frames acquired at time $0$ and $1$, and \textit{event sequence} $E_{0\rightarrow 1} $ consisting of events triggered during the time interval $t \in [0,1]$, and our goal is to insert one or more \textit{latent frame(s)} $\hat{I_{t}}$ at some time $t \in [0,1]$ between the key frames. Similar to previous works, we represent events as a \textit{voxel grid}~\cite{Zhu18eccvw}. We will use $V_{a\rightarrow b}$ to denote the voxel grid formed by converting events between times $t_a$ and $t_b$.

\begin{figure}[htb!]
    \centering
    \includegraphics[width=0.5\textwidth]{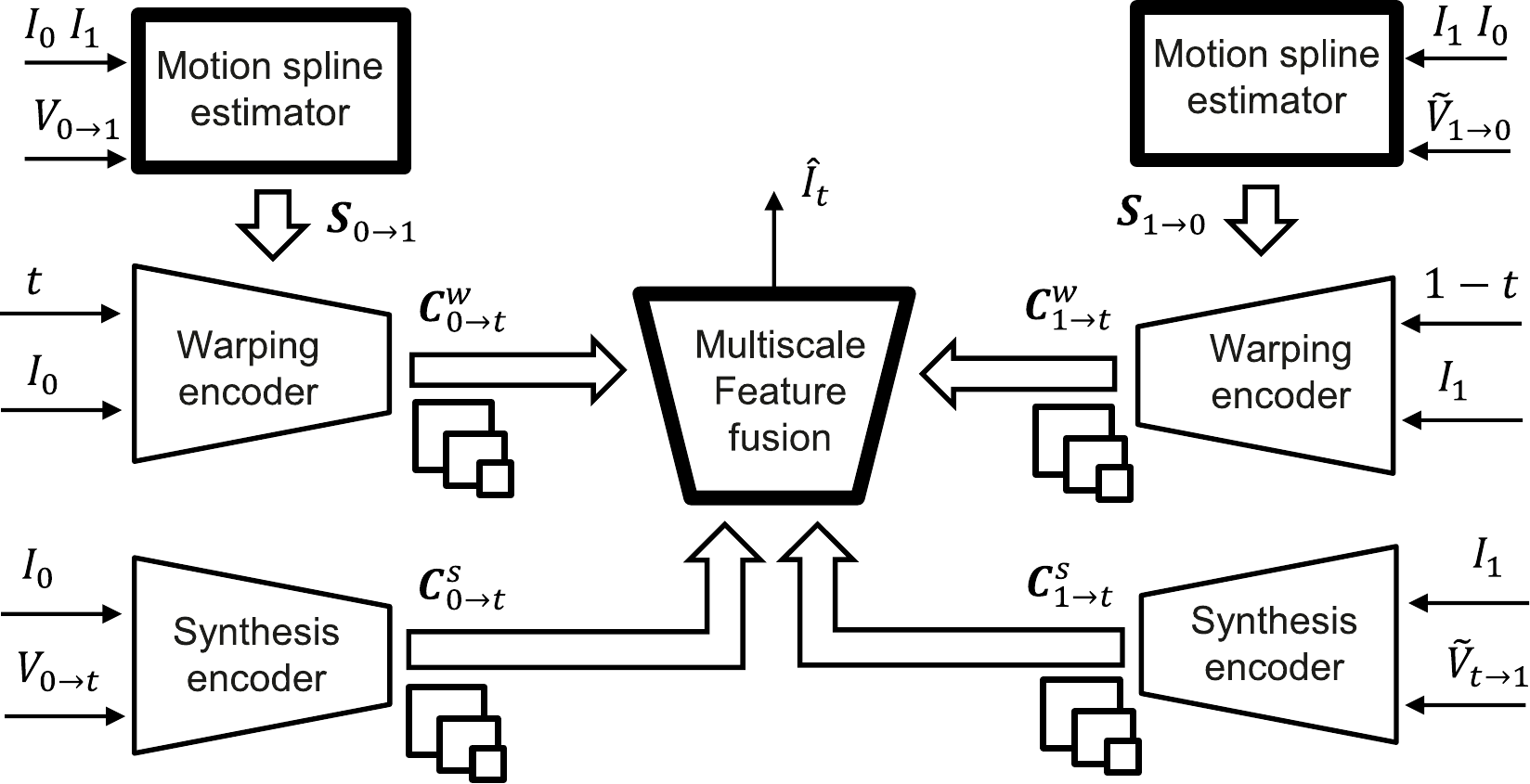}
    \caption{System overview. Main contribution of this work, shown with a thick contour, are: \textit{multi-scale feature fusion} that combines synthesis and warping-based interpolations on a feature level on multiple scales and \textit{motion spline estimation} that computes parametric motion model from boundary frames and inter-frame events. See ``system overview'' in  Section~\ref{sec:method} for more details.}
    \label{fig:system_overview}
\end{figure}

\textbf{System overview}. The overall system is shown in Figure~\ref{fig:system_overview} and key contributions are highlighted with a thick contour. Our system first generates multi-scale \textit{warping interpolation features} in two steps. First image $I_0$ is encoded using a warping encoder, resulting in multi-scale features. These features are then remapped to the time of the latent frame using splines $S_{0\rightarrow 1}$, derived from the voxel grid $V_{0\rightarrow 1}$ and images $I_0,I_1$, resulting in $\{\mathbf{C}^w_{0\rightarrow t}, \mathbf{C}^w_{1 \rightarrow t} \}$.
The \textit{warping interpolation features} are combined with \textit{synthesis interpolation features} $\{ \mathbf{C}^s_{0\rightarrow t}, \mathbf{C}^s_{1 \rightarrow t} \}$ computed from $I_0$ and  $I_1$, using a newly proposed \textit{multi-scale feature fusion} module and produces latent frame $\hat{I}_t$ at time $ t $. The system performs symmetric processing for $I_0$ and  $I_1$, therefore we only explain processing for $I_0$.   

The \textit{synthesis interpolation features} $\mathbf{C}^s_{0\rightarrow t}$ are computed by a \textit{synthesis encoder} from the preceding keyframe $ I_0 $ and voxel grid $ V_{0 \rightarrow t} $ for events between preceding and latent frames. Intuitively, this encoder ``adds'' intensity changes registered by events to the keyframe and thus can interpolate non-rigid objects with illumination changes, such as fire and water. In contrast to prior work, we found it beneficial to encode $I_0$ and $I_1$ separately using an encoder with shared weights, which we call \textit{shared synthesis encoder}.

The \textit{warping interpolation features} $\mathbf{C}^w_{0 \rightarrow t} $ are computed by a \textit{warping encoder} which warps features extracted from the preceding key frame $ I_0 $ to time $ t $ using pixel-wise \textit{spline} model of inter-frame motion $\mathbf{S}_{0\rightarrow 1} = \{ S^{\Delta x}_{0 \rightarrow t}, S^{\Delta y}_{0 \rightarrow t}, S^{p}_{0 \rightarrow t} \}$. 

The motion splines $\mathbf{S}_{0\rightarrow 1} $ are computed once per inter-frame interval using boundary frames $\{ I_0, I_1 \}$ and voxel grid $V_{0 \rightarrow 1}$ for inter-frame events using a newly introduced \textit{motion spline estimator}. Note, that in contrast to previous works, our motion estimator in addition to events uses boundary images to ensure interpolation robustness in low-contrast areas without events. In the next paragraphs, we explain the key modules of our system in detail.

\begin{figure*}[htb!]
    \centering
    \includegraphics[width=0.8\textwidth]{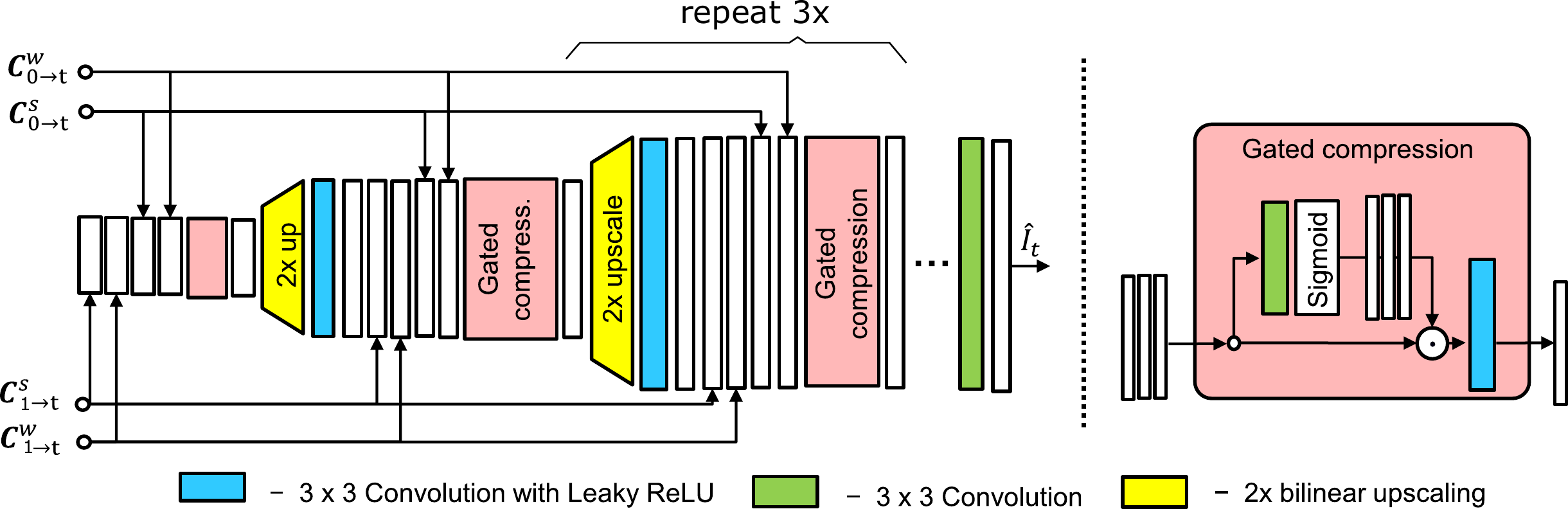}
    \caption{\textit{Multi-scale fusion} decoder progressively combines warping and synthesis interpolation features and features from the previous processing stage performed on a coarser scale using novel \textit{gated compression} module, that selects most informative features from each source. For detailed explanation please refer ``multi-scale feature fusion'' minisection in Section~\ref{sec:method}. Blank white boxes represents feature maps. $\bigodot$ represents the element-wise product. %
    }
    \label{fig:multiscale_fusion}
\end{figure*}

\textbf{Multi-scale feature fusion}. %
To improve fusion of synthesis and warping interpolation result, we use a \textit{multi-scale feature fusion decoder} shown in Figure~\ref{fig:multiscale_fusion}. The decoder progressively combines warping and synthesis interpolation features and features from the previous processing stage performed on a coarser scale. It relies on a novel \textit{gated compression} module, which attenuates features before combining them and thus, intuitively selects informative features from each source. For example, in Fig.~\ref{fig:gating}, the gated compression decides to use synthesis-based interpolation for the flame, for which optical flow fails (colder colors in image 1,
and warmer in image 2 of the figure). The gating idea was inspired by recent work on HDR fusion~\cite{yan_cvpr2019}, where it was used for combining multiple exposures. 

In~\cite{yu_iccv2021}, the authors also use a network with multiple encoders for combining images and events, however in contrast to their approach we \emph{(i)}~estimate motion not only from keyframes but also from events; \emph{(ii)}~move features of keyframes instead of keyframes; \emph{(iii)}~use gating instead of location-wise alpha blending for selecting most informative source; \emph{(iv)}~use not all inter-frame events in the synthesis encoder, but only events between keyframe and latent frame and \emph{(v)}~combine features from the previous coarse-scale processing stage with other features.
In the supplementary materials and ablations studies, we show the benefits of these design choices.         

\textbf{Motion spline estimator}. Previous work~\cite{tulyakov_gehrig_2021cvpr} computes motion independently from each latent frame to boundary frame by re-partitioning the events as shown in Figure~\ref{fig:motion_types}(c). This approach leverages information about non-linear inter-frame motion contained in events, however, its computational complexity scales linearly $O(N)$ with the number of interpolated frames $N$  and its motion estimates are independent and, thus, potentially temporally inconsistent. %

\begin{figure*}[htb]
    \centering
    \includegraphics[width=0.85\textwidth]{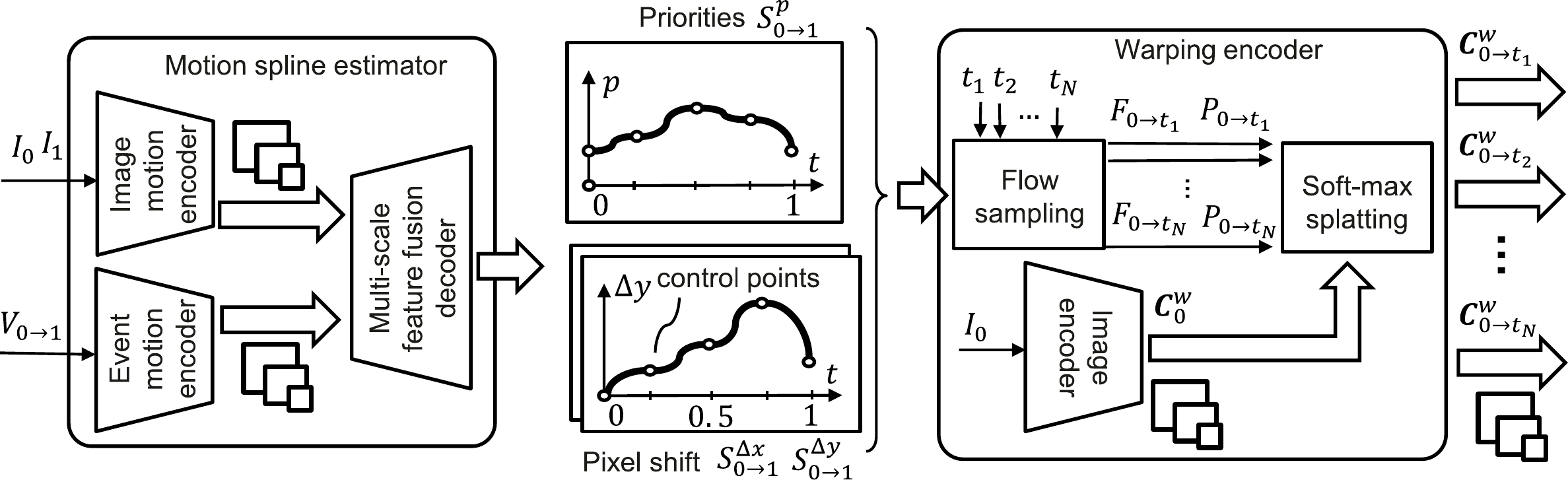}
    \caption{\textit{Motion spline estimator} uses boundary images in addition to inter-frame events to compute \textit{cubic motion splines} even in areas without events. Motion splines allow computing warping interpolation features for any inter-frame time, \textit{e.g.} $ \mathbf{C}^w_{0\rightarrow t_1}, \mathbf{C}^w_{0\rightarrow t_2}, \ldots \mathbf{C}^w_{0\rightarrow t_N} $ for time $ t_1, t_2, \ldots, t_N $ with minimum computational cost. For detailed explanation please refer to ``motion spline estimator'' in Section~\ref{sec:method}.}
    \label{fig:motion_estimator}
    \vspace{-2ex}
\end{figure*}

In contrast, this work relies on a \textit{spline motion estimator} shown in Figure~\ref{fig:motion_estimator} which infers nonlinear inter-frame motion~(see Figure~\ref{fig:motion_types}(d)) from events and key frames and approximates it with splines. This enables efficient sampling of motion between key frames and arbitrary latent frame and ensures temporal consistency of the motion. Our motion estimator computes 3 cubic splines  $ \{ S^{\Delta x}_{0 \rightarrow 1}, S^{\Delta y}_{0 \rightarrow 1}, S^{p}_{0 \rightarrow 1} \}$ for each location. These splines model horizontal, vertical displacement and warping priority of each pixel of the preceding key frame as a function of time, correspondingly. Each spline we represent by $K$ control points, \textit{e.g.}, horizontal displacement spline $S^{\Delta x}_{0 \rightarrow 1}$ we represent as displacements $ (\Delta x_0, \Delta x_{\frac{1}{K-1}}, \Delta x_{\frac{2}{K-1}}, \ldots, \Delta x_1)$ for uniformly sampled timestamps $ (0, \frac{1}{K-1}, \frac{2}{K-1}, \ldots, 1) $. 

Using these splines, we can compute multi-scale warping interpolation features $ \mathbf{C}^w_{0\rightarrow t} $ for any time $ t \in [0, 1] $ with minimum additional computational cost using the \textit{warping encoder}. To this end, for a given time $t$ the \textit{flow sampling} module firstly samples flow $F_{0\rightarrow t}$ and priority $P_{0 \rightarrow t}$ from the splines using the~\textit{cubic convolution method}~\cite{keys_tassp1981}. Then, using the flow and the priority, the \textit{softmax splatting} module projects multi-scale image features $ \mathbf{C}^w_{0} $, extracted from the preceding boundary image $I_0$ by \textit{image encoder} to time $t$. Note, that we experimented with different representations of inter-frame motion, including polynomials, Bezier splines, Natural B-Splines and they all performed very similar. We adopted the representation described above since cubic convolution methods allow to sample motion efficiently.  

\textbf{Using images and events for motion estimation}: Time lens only uses events for motion estimation because they contain information about non-linear inter-frame motion. However, in low contrast areas, where temporal brightness changes are below the contrast threshold of a camera, events are not triggered~\cite{gallego_pami2020}. This causes interpolation artifacts such as shown in Figure~\ref{fig:eyecatcher}(c). These artifacts could be avoided if Time Lens used images in addition to events. However, since it relies on bi-linear interpolation~\cite{jaderberg_2015} for image warping, it requires motion from the ``non-existing'' latent frame to boundary frames shown in Figure~\ref{fig:motion_types}(c), which can only be approximated from the keyframes. 

We use \textit{softmax splatting}~\cite{niklaus_2020} interpolation for warping which requires motion from boundary frame to latent frame (see  Figure~\ref{fig:motion_types}(d)) and, thus, allows combining events and image-based motion estimation in the proposed \textit{motion spline estimator}. This motion  estimator has two encoders: \textit{image-based motion encoder} for estimating motion features from the boundary images $\{I_0, I_1\}$, and \textit{event-based motion encoder} for estimating motion features from voxel grid $ V_{0 \rightarrow 1} $ of inter-frame events. These motion features are then combined in the joint decoder similar to  Figure~\ref{fig:multiscale_fusion}. We use an architecture with two encoders, because it is easier for the network to learn motion from events than images, and therefore in presence of a single encoder network simply converges to a local minimum and mostly ignores images.

\section{Experiments}
\label{sec:experiments}
Next, we justify our design for multi-scale fusion and spline motion estimation modules with a series of ablation studies. Then, we compare our VFI method to other state-of-the-art image and event-based methods on several benchmark datasets, including our newly introduced \textit{Beam Splitter Events \& RGB~(BS-ERGB)} dataset.

All experiments are done using the \textit{PyTorch} framework~\cite{pytorchurl}. We use the \mbox{Adam} optimizer\cite{Kingma15iclr}, batch size 4 and learning rate $10^{-4}$, which we decrease by a factor of $10$ every 12 epochs. We train each module for 27 epochs. For training, we use a large-scale dataset with synthetic events generated from the \textit{Vimeo90k} septuplet dataset~\cite{xue_2019} using the video to events conversion method~\cite{gehrig_2020}. 

We train the motion spline and multi-scale fusion module separately and then fine-tune them together. Firstly, we train the motion module with $L_1$ and SSIM losses with weights 0.15 and 0.85. Then, we freeze the motion network and train the fusion network with LPIPS~\cite{zhang_2018} and $L^1$ losses with weights 1.0 and 2.0. For training each module we also use multistage training, which firstly trains each encoder separately with dummy decoders and then freezes encoders and training decoders, and finally trains the module. This training procedure significantly boosts the performance of our method. Due to the time limitations, we don't use this method in ablations, if not stated explicitly. For real data, we fine-tune our entire network with the losses that we use for training the fusion module. To measure the quality of interpolated images we use peak signal-to-noise ratio~(PSNR) and structural similarity~(SSIM)~\cite{wang_2004} metrics and LPIPS. 

\begin{table}[]
\centering
\resizebox{1\linewidth}{!}{%
\begin{tabular}{l|cc|cc|cc}
\hline
\textbf{Method} & \multicolumn{4}{c|}{\textbf{BS-ERGB}} & \multicolumn{2}{c}{\textbf{HS-ERGB\cite{tulyakov_gehrig_2021cvpr} }} \\
 & \multicolumn{2}{c|}{\textbf{1 skip}} & \multicolumn{2}{c|}{\textbf{3 skips}} & \multicolumn{2}{c}{\textbf{7 skips}}\\
\textbf{} & PSNR~$\uparrow$ & LPIPS~$\downarrow$ & PSNR~$\uparrow$ & LPIPS~$\downarrow$ & PSNR~$\uparrow$ & LPIPS~$\downarrow$\\ \hline%
FLAVR\cite{kalluri_arxiv2021} & 25.95 & 0.086 & 20.90 & 0.151 & 27.42 & 0.031 \\
DAIN\cite{bao_cvpr2019} & 25.20 & 0.067 & 21.40 & 0.113 & 29.82 & 0.022 \\
Super Slowmo~\cite{jiang_2018} & - & - & 22.48 & 0.115 & 30.05 & 0.103\\
QVI~\cite{xu_2019} & - & - & 23.20 & 0.110 & 26.28 & 0.143\\
Time Lens\cite{tulyakov_gehrig_2021cvpr} & 28.36 & 0.026 & 27.58 & 0.031 & \textbf{33.48} & 0.017 \\
\textbf{Ours} & \textbf{28.56} & \textbf{0.0222} & \textbf{27.63} & \textbf{0.026} & 33.09 & \textbf{0.016}\\ \hline%
\end{tabular}}\caption{Quantitative comparison of our method with frame- and hybrid frame+event-based methods in terms of PSNR (higher is better) and LPIPS (lower is better). For on the HS-ERGB dataset\cite{tulyakov_gehrig_2021cvpr} we average the score over the \emph{close} and \emph{far} subsets.}\label{tab:benchmarking}
\end{table}
\subsection{Ablation studies}
Next, we ablate the various components of our method. For ablation studies, we randomly sample 4k training examples and 500 validation examples from the Vimeo90k~\cite{xue_2019} dataset. We first ablate the motion estimation module, and then the fusion module. For the motion module we calculate errors in non-occluded areas of warped frames. For fusion, we compute the error of final interpolated frames. Results are summarized in Table~\ref{tab:ablations:warping} for the motion module, and in Table~\ref{tab:ablations:fusion} for the fusion module.

\begin{table}[h]
    \small
    \centering
    \resizebox{0.8\linewidth}{!}{%
\begin{tabular}{lcc}
\hline
\multicolumn{1}{l|}{\textbf{Method} \hfill \hspace{0.5cm} \textbf{15 frames[ms]}}                   & \textbf{SSIM}  & \textbf{PSNR [dB]} \\ \hline
\multicolumn{3}{c}{\textit{Importance of Spline Flow}}                                       \\ \hline
\multicolumn{1}{l|}{Linear \hfill  \hspace{2.5cm} 200}                            & 0.856          & 26.83              \\
\multicolumn{1}{l|}{Non-parametric \hfill  \hspace{1.2cm} 2700}                      & \textbf{0.877} & \textbf{28.20}     \\
\multicolumn{1}{l|}{\textbf{Spline (ours)} \hfill  \hspace{1.6cm}  \textbf{220}}            & 0.863          & 27.41              \\ \hline
\multicolumn{3}{c}{\textit{Importance of Images}}                                            \\ \hline
\multicolumn{1}{l|}{Images}                            & 0.808          & 24.51              \\
\multicolumn{1}{l|}{Events}                            & 0.853          & 26.82              \\
\multicolumn{1}{l|}{\textbf{Images and Events (ours)}} & \textbf{0.863} & \textbf{27.41}     \\ \hline
\multicolumn{3}{c}{\textit{Comparison with State-of-the-art}}                                \\ \hline
\multicolumn{1}{l|}{EV-FlowNet}                        & 0.756             & 22.31                 \\
\multicolumn{1}{l|}{Time Lens Flow}                    & 0.866          & 27.22              \\
\multicolumn{1}{l|}{\textbf{Ours}}                     & \textbf{0.879} & \textbf{28.10}     \\ \hline
\end{tabular}}
    \caption{Ablation for motion estimation module: while spline motion benefits run-time performance for multiple frame interpolation, using images and events boosts accuracy. %
    Compared to methods Time lens flow~\cite{tulyakov_gehrig_2021cvpr} and EV-FlowNet~\cite{Zhu18rss}, our method achieves superior performance.}
    \label{tab:ablations:warping}
\end{table}
\begin{table}[h]
    \centering
    \footnotesize
    
\begin{tabular}{lcc}
\hline
\multicolumn{1}{l|}{\textbf{Method}}                       & \textbf{SSIM}  & \textbf{PSNR [dB]} \\ \hline
\multicolumn{3}{c}{\textit{Importance of Fusion}}                                                \\ \hline
\multicolumn{1}{l|}{Warping}                               & 0.886          & 29.42              \\
\multicolumn{1}{l|}{Synthesis}                             & 0.868          & 29.77              \\
\multicolumn{1}{l|}{\textbf{Synth. \& warping (ours)}} & \textbf{0.912} & \textbf{31.87}     \\ \hline
\multicolumn{3}{c}{\textit{Importance of Gating}}                                                \\ \hline
\multicolumn{1}{l|}{No gating}                             & 0.907          & 31.67              \\
\multicolumn{1}{l|}{\textbf{Gating (ours)}}                & \textbf{0.912} & \textbf{31.87}     \\ \hline
\multicolumn{3}{c}{\textit{Comparison with State-of-the-art}}                                    \\ \hline
\multicolumn{1}{l|}{Time Lens Fusion}                      & 0.906          & 31.25              \\
\multicolumn{1}{l|}{\textbf{Ours}}                         & \textbf{0.919} & \textbf{32.73}     \\ \hline
\end{tabular}
    \caption{Ablation studies for fusion module: combining synthesis and warping features boosts performance, as does gated compression mechanism. Compared to image-level fusion in \cite{tulyakov_gehrig_2021cvpr}, our method achieves superior performance.}
    \label{tab:ablations:fusion}
\end{table}

\begin{figure}[htbp]
    \begin{subfigure}[b]{0.245\columnwidth}
	\includegraphics[width=1\columnwidth,trim={0 0.15cm 0 0},clip]{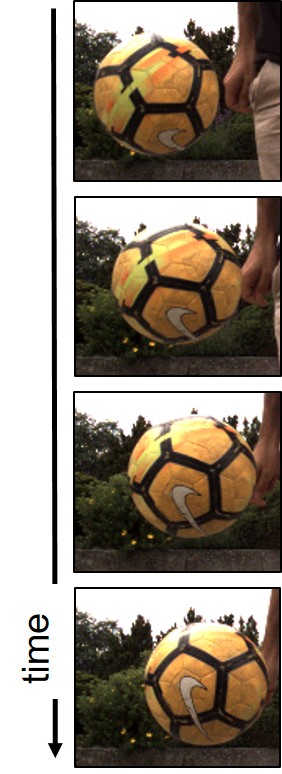}
	\end{subfigure}
	\begin{subfigure}[b]{0.32\columnwidth}
	\animategraphics[width=1\columnwidth,autoplay,loop]{5}{figures/nonlinear_flow/}{0}{3}
	\end{subfigure}
	\;
    \begin{subfigure}[b]{0.38\columnwidth}
	\includegraphics[width=1\columnwidth,trim={0 0.15cm 0 0},clip]{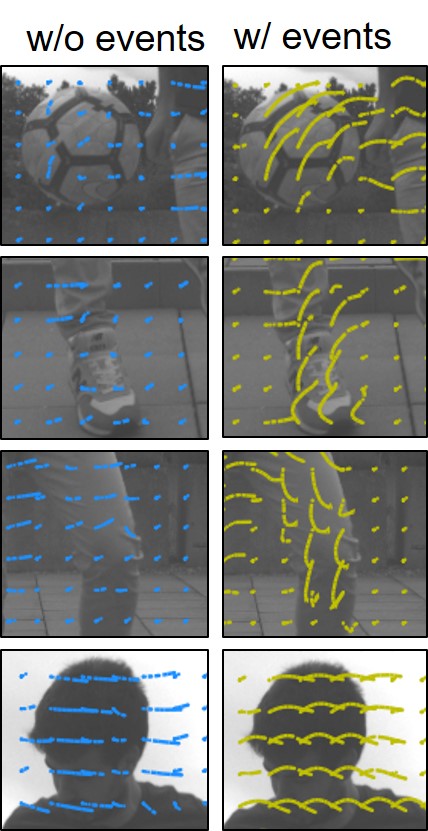}
	\end{subfigure}
\caption{Spline motion visualization. The left image shows ground truth image sequence with close-ups~(animation can be viewed in Adobe). The right image shows motion estimated by proposed motion spline estimator from images \& events~(green) and from images only~(blue).}
	\label{fig:nonlinear_flow}
\end{figure}

\subsubsection{Motion Estimation Module}

\emph{Visualization of Spline Motion}. We visualize the output of our spline estimator in Figure~\ref{fig:nonlinear_flow} (right, in yellow). By combining images with events, it can model the highly non-linear trajectory of a soccer ball. 
Events are an integral part of modeling non-linearity since when they are removed (blue, right), the flow module defaults to using images alone and predicts linear motion. 

\emph{Importance of Spline Motion.} We compare against linear motion and non-parametric motion in Table~\ref{tab:ablations:warping}. The results suggest that non-parametric motion estimation achieves higher accuracy (28.20 dB vs. 27.41 dB) but has a much higher run time (2700 ms vs. 220 ms) for 15 flow predictions since it needs to run for each inserted frame, while spline motion can be efficiently re-sampled once computed. This difference becomes especially noticeable when inserting multiple frames. By contrast, linear motion estimation has a low run time but also low performance.

\emph{Importance of Images}. We train two variations of our motion module: one using only events, and one using only frames and report results in Table~\ref{tab:ablations:warping}. We note that combining both sensor inputs achieves the best results, boosting performance by 0.6 dB or 2.9 dB compared to single-sensor inputs. 
This underlines the complementarity of their information: while events provide non-linear motion cues, images provide information where events are missing. %

\emph{Comparison to State-of-the-art}. We compare against optical flow methods EV-FlowNet~\cite{Zhu18rss} and the optical flow module from Time Lens\cite{tulyakov_gehrig_2021cvpr}, which both predict non-parametric flow only from events. The resulting warping error in terms of PSNR is in Table~\ref{tab:ablations:warping}. 
Our method outperforms the runner up \cite{tulyakov_gehrig_2021cvpr} by 0.88 dB in terms of PSNR. Note that here we use multistage training  explained in Section~\ref{sec:experiments}.

\begin{figure}[h!]
    \hspace{1em}%
    \begin{subfigure}[b]{0.98\columnwidth}
	\includegraphics[width=1\columnwidth,trim={0 0 0 0},clip]{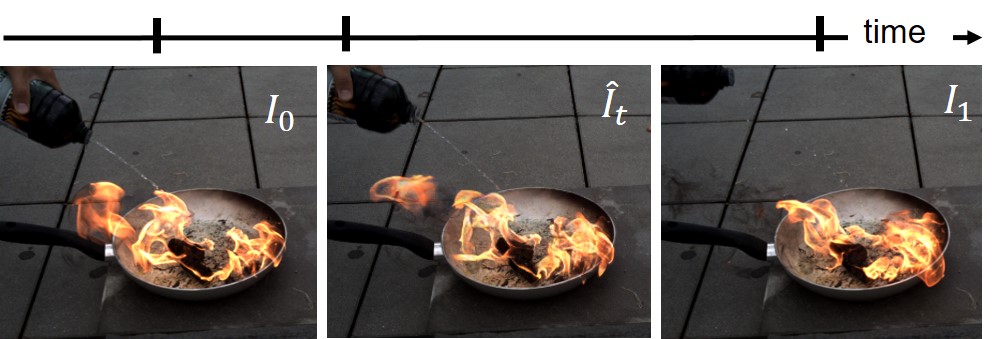}
	\end{subfigure}\\
	\begin{subfigure}[b]{1\columnwidth}
	\includegraphics[width=1\columnwidth,trim={0.2cm 0 0 0},clip]{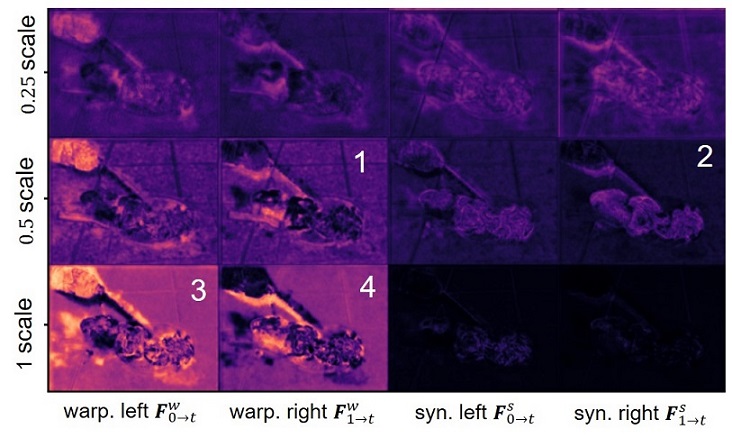}
	\end{subfigure}%
\caption{Gated compression weights visualization. The top figure shows key frames and latent interpolated frame. The bottom figure shows average weight prediction by gated compression for synthesis and warping features on each scale (smaller weight shown in colder colors).%
For details, please refer to ``gated compression weight'' mini-section in ~Section~\ref{sec:fusion_ablation}}
	\label{fig:gating}
	\vspace{-2ex}
\end{figure}

\begin{figure*}[h!]
	\centering
    \animategraphics[width=0.22\textwidth,autoplay,loop]{5}{figures/qualitative/acquarium_0}{2}{6}%
	\includegraphics[width=0.27\textwidth]{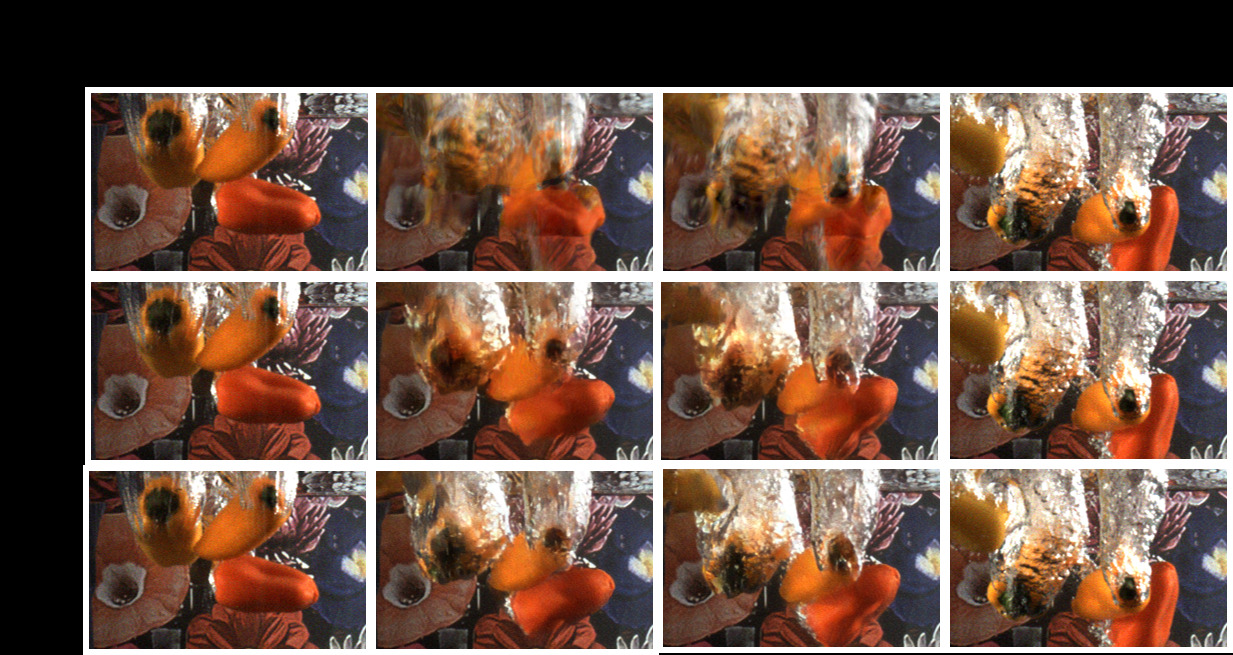} \,%
	\animategraphics[width=0.22\textwidth,autoplay,loop]{5}{figures/qualitative/axing_}{0}{5}%
	\includegraphics[width=0.278\textwidth]{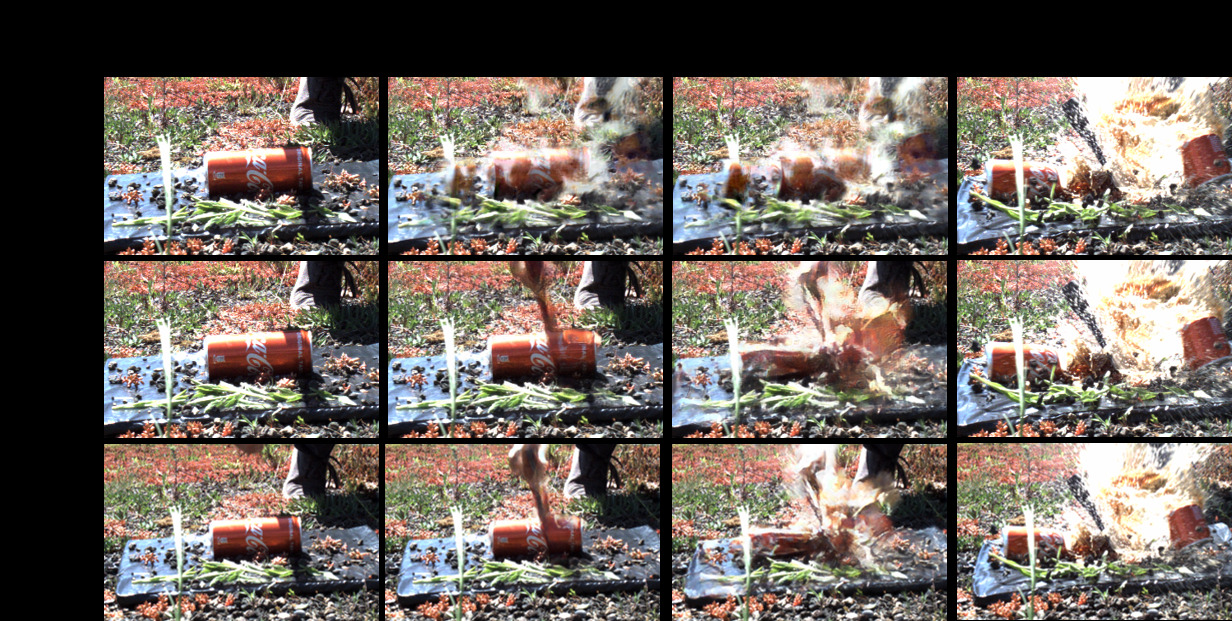}
	\vspace{-0.4cm}\\
	\animategraphics[width=0.2\textwidth,autoplay,loop]{5}{figures/qualitative/fire_}{0}{5}%
	\includegraphics[width=0.29\textwidth]{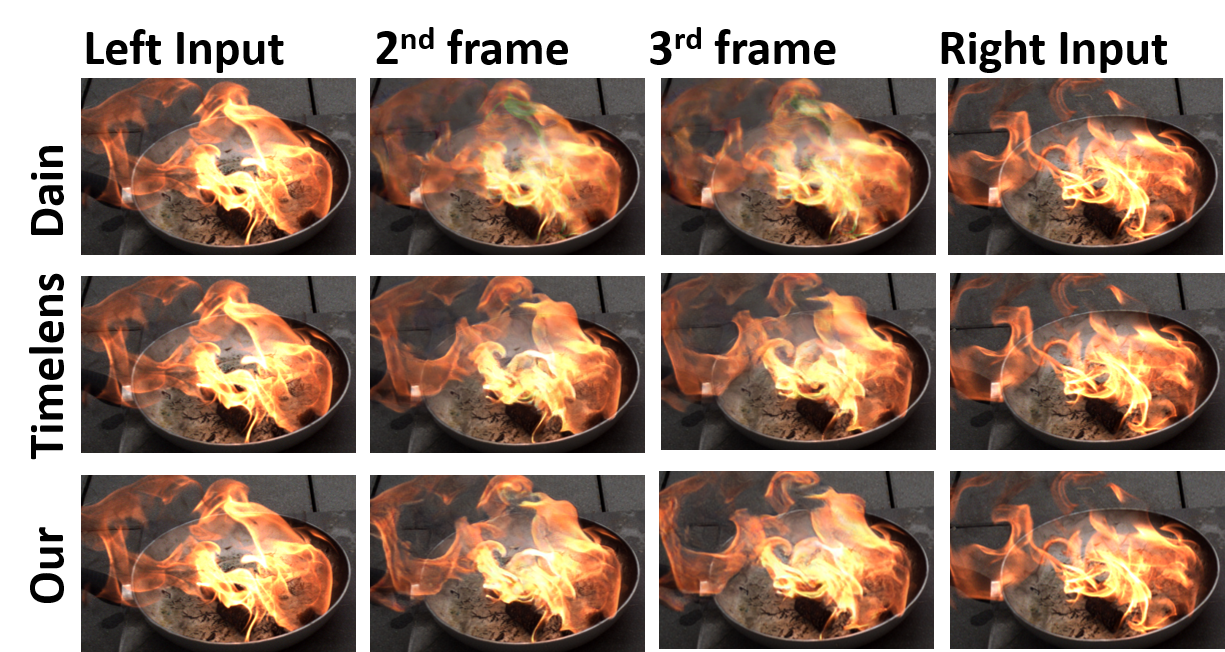} \,%
    \animategraphics[width=0.20\textwidth,autoplay,loop]{5}{figures/qualitative/pen_0}{1}{6}%
	\includegraphics[width=0.30\textwidth,trim={0 0 0.7cm 0},clip]{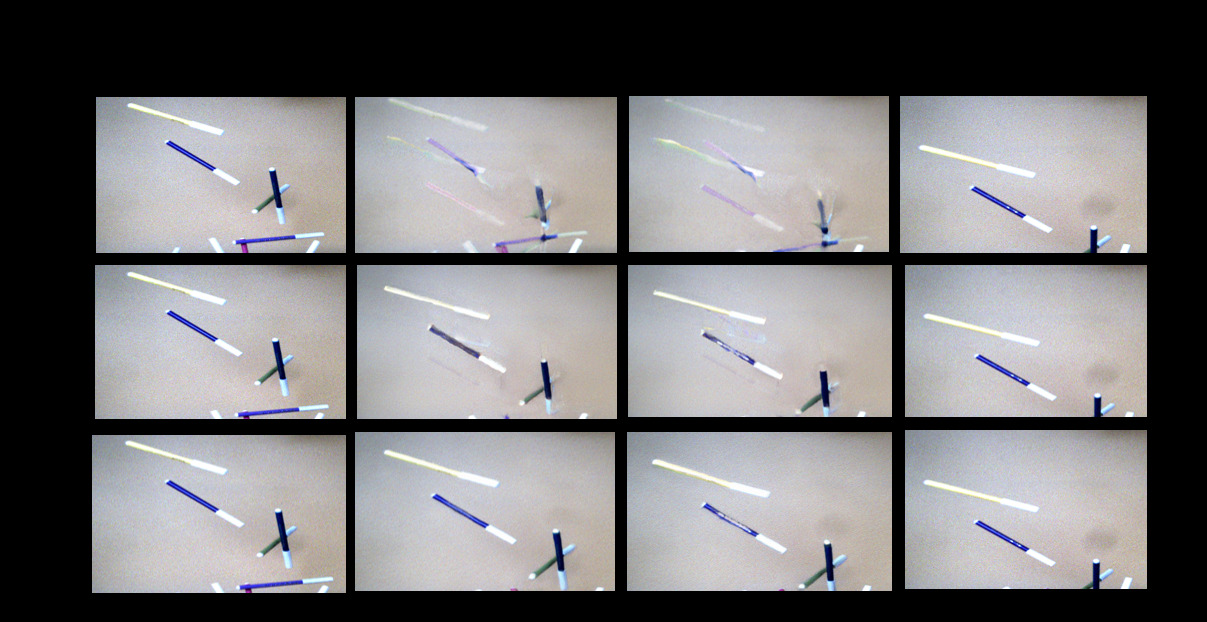}\\

\caption{Qualitative results for the proposed method and its closes competitor on test sequences from the BS-ERGB: ``Water''~(top-left), ``Fire''~(bottom-left) ``Fast motion''~(top-right) and ``Thin objects"~(bottom-right). For each sequence, the figure shows our interpolation results~(the animation can be viewed in Acrobat Reader) and close-up interpolation results on the right. }
\vspace{-2ex}
	\label{fig:qualitative}
\end{figure*}

\subsubsection{Multi-scale Fusion}\label{sec:fusion_ablation}

\emph{Gated Compression.} We firstly confirm the importance of gated compression by training the fusion network without gated compression. As shown in  Table~\ref{tab:ablations:fusion}~(\emph{Importance of Gating}), gated compression improves the fusion module. Next, in Figure~\ref{fig:gating} we show channel-wise averaged weights predicted by the gated compression module shown in Figure~\ref{fig:multiscale_fusion} for synthesis and warping features on each scale for a specific example. We conclude that: \emph{(i)} synthesis features are used for non-rigid objects, such as fire, while warping features are used for rigid objects, such as the bottle (compare 1 \& 2); 
\emph{(ii)} areas occluded in the closer left frame $I_0$ are filled from the right frame $I_1$ (e.g. see 3 \& 4); \emph{(iii)} on a finer scale warping interpolation features are preferred to synthesis features. %

\emph{Importance of Fusion:} We first study the effect of combining synthesis and warping features during fusion, which we show in Table~\ref{tab:ablations:fusion}. We see that combining features from both modules boosts performance by 2.1 dB.  

\emph{Comparison to State-of-the-art:} We also compare against the image-level fusion module from~\cite{tulyakov_gehrig_2021cvpr}. As shown in Table~\ref{tab:ablations:fusion} multi-scale feature-level fusion performs better by 1.48 dB. Note that for this final comparison we use multistage training explained in Section~\ref{sec:experiments}.

\begin{table}[tb]
    \caption{Details of proposed BS-ERGB dataset compared to similar GEF dataset.~\cite{wang_2020b}}\label{tab:bsergb_summary}
    \vspace{0.5em}
    \centering
    \small
    \begin{tabular}{lll}
    \hline
    &  \textbf{BS-ERGB~Ours)} & \textbf{GEF}\cite{wang_2020b} \\
    \hline
    Event Camera  &  $\mathbf{970\times625}$ & $190\times180$ \\
    \textnumero~sequences  & \textbf{123} & 20 \\
    Scene dynamic & \textbf{high-speed} & low-speed \\ 
    RGB Camera  &  $970\times625$, 28~fps & $1520\times1440$, 20~fps \\
    Camera motion & moving \& static & moving \& static \\
    Seq. length & 100-600 frames & 200-250 frames \\
     \hline
    \end{tabular}
\end{table}

\subsubsection{Beamsplitter Events and RGB dataset}\label{sec:dataset}

We built a new hybrid setup, that uses a FLIR 4096$\times$2196 RGB global shutter camera and a Prophesee Gen4 1280$\times$720 Event camera mounted on a rigid case with a 50/50 one-way mirror to share incoming light. %
With this setup, we have collected BS-ERGB dataset, summarized in Tab.~\ref{tab:bsergb_summary}. As shown in the table, compared to the GEF dataset~\cite{wang_2020b}, proposed dataset is larger, has higher event resolution and contains high-speed scenes, with a moving and static camera. Also, in contrast to GEF, the proposed datasets include challenging scenarios with fire, water, shadows, transparent objects, high-frequency patterns, small and colorful objects, thin structures, close and far away scenes, capturing both linear and highly non-linear motion. 

\subsubsection{Benchmarking}

We quantitatively compare our method to state-of-the-art image-based and image-and-event-based VFI methods HS-ERGB~\cite{tulyakov_gehrig_2021cvpr} and newly introduced BS-ERGB dataset, featuring realistic events and high-quality images. As shown in Table~\ref{tab:benchmarking}, the proposed method greatly outperforms not only the best image-based methods, but also the event-and-image-based Time Lens by up to 0.2 dB in PSNR and improves its LPIPS score by up to 15\%.     

In Figure~\ref{fig:qualitative} we also show qualitative comparison of our method to state-of-the-art image-based  DAIN~\cite{bao_cvpr2019} and image-and-event-based Time Lens~\cite{tulyakov_gehrig_2021cvpr} methods. From the examples, it is clear that our method outperforms DAIN on the scenes with deformable objects and illumination change, such as scenes with fire and water splashes. The proposed method also outperforms Time Lens on especially complex scenes with
thin and low contrast objects. 

\rev{\textbf{Timing Experiments:} We highlight the computational efficiency of our method by comparing the per-frame computation time at different upsampling factors, for both TimeLens~\cite{tulyakov_gehrig_2021cvpr} and our method. The results are reported in Tab.~\ref{tab:timing}. Especially at high upsampling factors ($>10$) our method uses less computation per frame, leveraging the continuous flow, while TimeLens requires a constant computation time. 

\begin{table}[h]
\begin{tabular}{lccccc}
\hline
\multirow{2}{*}{} & \multicolumn{4}{c}{\textbf{Time per frame~[ms]}} & \textbf{\textnumero~param} \\
Upscale Factor& $1\times$ & $3\times$ & $10\times$ & $20\times$ & \textbf{[M]}\\
\hline
Ours & 634 & 303 & \textbf{193} & \textbf{167} & \textbf{53.9} \\
Time Lens~[22]  & \textbf{273} & \textbf{273} & 273 & 273 & 72.2 \\
\hline
\end{tabular}
\caption{Per-frame computation at different upsampling factors.}
\label{tab:timing}
\end{table}
}

\section{Conclusion}
Producing stunning slow-motion video from a low framerate video has been recently made possible through the advent of frame-based and event-based methods. Event-based methods, in particular, manage to capture the non-linear motion between keyframes, especially in high-speed and highly dynamic scenarios, by using events in the blind-time between frames. 
Existing event-based approaches rely on simple fusion and chunked flow which causes ghosting, temporal inconsistencies, and high computational latencies. In this work, we presented a novel method that addresses all of these limitations. By leveraging continuous flow, our method produces a temporally consistent flow that can be efficiently queried at several timesteps for multi-frame insertion. Finally, our novel multi-scale fusion module significantly reduces ghosting artifacts. We demonstrate an up to 0.2 dB improvement in PSNR and 15\% improvement in LPIPS score over state-of-the-art methods while being significantly faster.

\section{Acknowledgment}
This work was supported by Huawei, and as a part of NCCR Robotics, a National Centre of Competence in Research, funded by the Swiss National Science Foundation (grant number 51NF40\_185543).

\section{Dataset Licenses}
In this work we use the HS-ERGB dataset [22] which is published under the "TimeLens Evaluation License" and can be found under the URL \url{http://rpg.ifi.uzh.ch/timelens}. The Vimeo90k dataset [26] is published under the following URL \url{http://toflow.csail.mit.edu/}.

\section{Network structure}
\begin{figure}[htb!]
    \centering
    \includegraphics[width=0.5\textwidth]{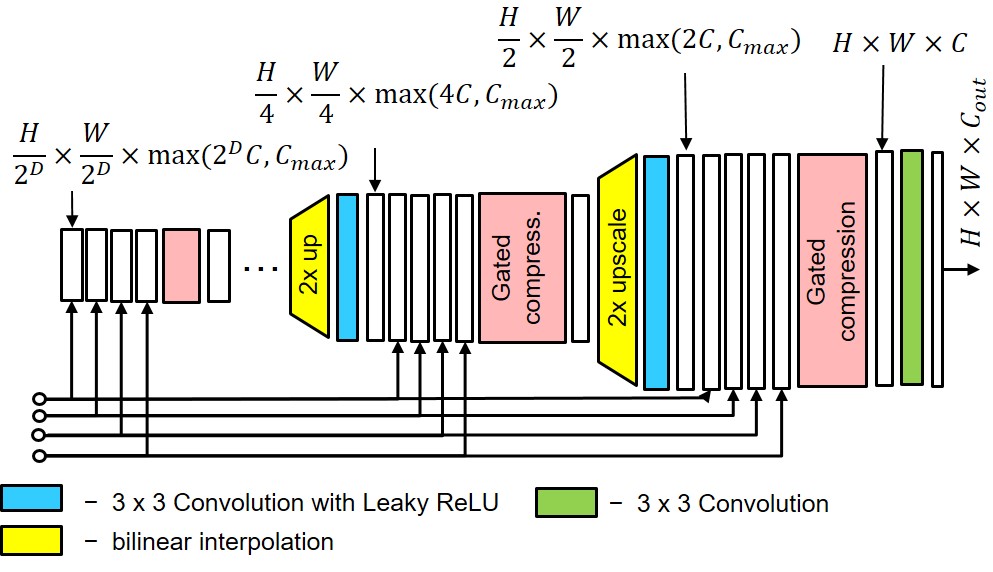}
    \caption{Architecture of joint decoder.}\label{fig:decoder}%
    \vspace{0.3cm}
    \includegraphics[width=0.5\textwidth]{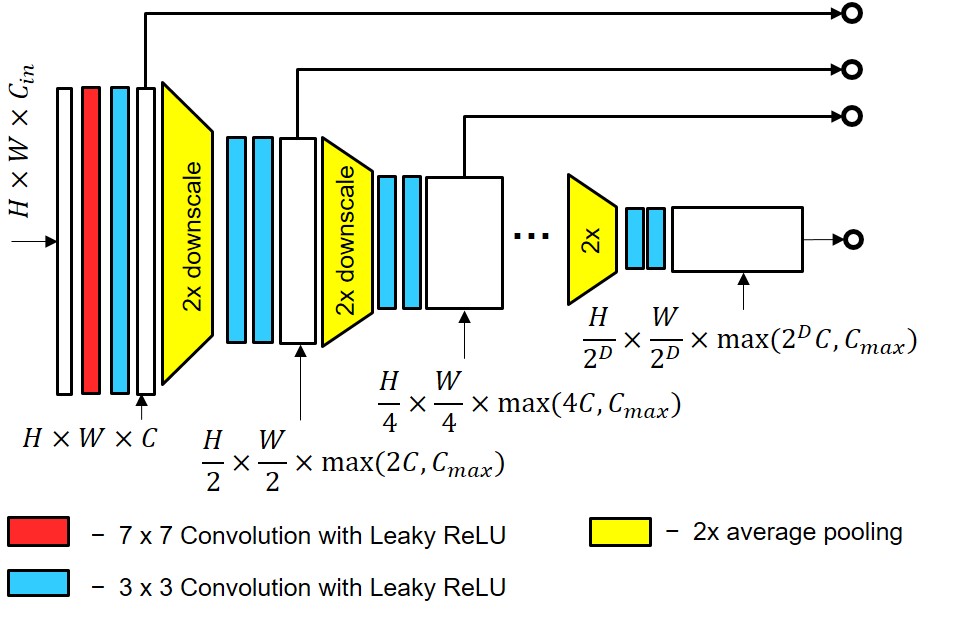}
    \caption{Architecture overview of encoder.}\label{fig:encoder}
\end{figure}

In this section, we provide a detailed description of our network. As explained in the paper, our network consist of \textit{spline motion estimator} and \textit{multi-scale feature fusion} modules. The spline motion estimator consists of 2 encoders and one joint decoder and the multi-scale fusion network consists of 4 encoders and one joint decoder. In both networks, encoders and decoders have the same structure shown in Fig.~\ref{fig:encoder} and Fig.~\ref{fig:decoder}. 

All decoder and encoder are defined by three parameters: depth $D$, the maximum number of features $C_{max}$ and the base number of features $ C $. We provide these parameters for the motion estimator module and multi-scale fusion module in Tab.~\ref{tab:encoder_and_decoder_parameters}.

\begin{table}[h]
    \centering
    \small
    \begin{tabular}{llll}
    \textbf{Module} &  $D$ & $ C_{max} $ &  $ C $ \\
    \hline
    \textbf{Fusion encoders \& decoder} & 3 & 128 & 32 \\
    \textbf{Motion estimator encoders \& decoder} & 4 & 256 & 64 \\
    \hline
    \end{tabular}
    \caption{Parameters of encoder and decoder for motion estimator and multi-scale fusion modules.}
    \label{tab:encoder_and_decoder_parameters}
\end{table}

\section{Training Details}
In this section, we provide additional details about the training procedure.

\textit{Thresholded losses}. As we mentioned in the paper, we use SSIM and $L^1$ losses to train the motion estimator. We found that it is beneficial to use these losses only in the areas where they are beyond a certain threshold~(0.06 for $L^1$ and 0.4 for SSIM). Intuitively, this helps to exclude occluded areas and areas with brightness constancy violation from the loss computation.  

\textit{Multi-stage training}. We developed a multi-stage training procedure that improves the performance of the motion estimator~(see~Tab.~\ref{tab:additional_ablation:motion}) as well as the multi-scale fusion module~(see Tab.~\ref{tab:additional_ablations:fusion}). We use the same procedure for the motion encoder and the multi-stage fusion module since they share a similar high-level network structure. Firstly, we train the image and event-based encoders separately, each with its own ``dummy'' decoder. Secondly, we select similar performing checkpoints for each encoder, we remove the dummy encoders and train joint decoder while freezing parameters of the encoders. Finally, we unfreeze all weights and train the entire network. This procedure helps when encoders have different training speeds and forces the network to avoid a local minimum where the decoder uses features only from the ``faster'' training encoder, ignoring the other. For example, the event-based motion encoder in the motion estimation module trains much faster than the image-based motion encoder, and thus using a single-stage training procedure would learn to simply ignore images and only use events. By contrast, our multi-stage training procedure forces the network to use both inputs.

\section{Additional Ablations}\label{sec:additional_ablations}
Here we present ablations that we omitted from the main paper due to space limitations.

\begin{table}[h!]
    \small
    \centering
\begin{tabularx}{0.5\textwidth}{LCC}
\hline
\multicolumn{1}{l|}{\textbf{Method} \hfill \hspace{0.5cm}}                   & \textbf{SSIM}  & \textbf{PSNR [dB]} \\ \hline
\multicolumn{3}{c}{\textit{Importance of Double Encoder}}                                    \\ \hline
\multicolumn{1}{l|}{U-Net}                             & 0.858          & 27.13              \\
\multicolumn{1}{l|}{\textbf{Double Enc. U-Net (ours)}} & \textbf{0.863} & \textbf{27.41}     \\ \hline
\multicolumn{3}{c}{\textit{Importance of Multistage Training}}                               \\ \hline
\multicolumn{1}{l|}{Single stage}                      & 0.863          & 27.41              \\
\multicolumn{1}{l|}{\textbf{Multistage (ours)}}        & \textbf{0.879} & \textbf{28.10}     \\ \hline
\end{tabularx}%
\caption{Additional ablation for motion estimation module.}\label{tab:additional_ablation:motion}
\vspace{0.3cm}    
\begin{tabularx}{0.5\textwidth}{LCC}
\hline
\multicolumn{1}{l|}{\textbf{Method}}                       & \textbf{SSIM}  & \textbf{PSNR [dB]} \\ \hline
 \multicolumn{3}{c}{\textit{Importance of shared synthesis encoder}}                                                   \\ \hline
 \multicolumn{1}{l|}{Joined}                                & 0.911          & 31.74              \\
 \multicolumn{1}{l|}{\textbf{Shared (ours)}}                & \textbf{0.912} & \textbf{31.87}     \\ \hline
 \multicolumn{3}{c}{\textit{Importance of Multistage Training}}                                   \\ \hline
 \multicolumn{1}{l|}{Single stage}                          & 0.912          & 31.87              \\
 \multicolumn{1}{l|}{\textbf{Multistage (ours)}}            & \textbf{0.919} & \textbf{32.73}     \\ \hline
\end{tabularx}
    \caption{Additional ablation studies for fusion module.}\label{tab:additional_ablations:fusion}
\end{table}

\emph{Motion estimator}. Here we summarize additional ablations for the motion estimator in Table~\ref{tab:additional_ablation:motion}. We found that using two separate encoders in the motion module improves results by 0.23 dB, by leaving the encoders more freedom to compute separate features for events and frames. Additionally, we found that by conducting multi-stage training, we can further improve the performance of our module by 0.69 dB. Multistage training consists of firstly training each encoder separately, when freezing encoders and train decoder, and finally training the entire motion module. 

\begin{figure}[h!]
    \centering
    \includegraphics[width=0.35\textwidth]{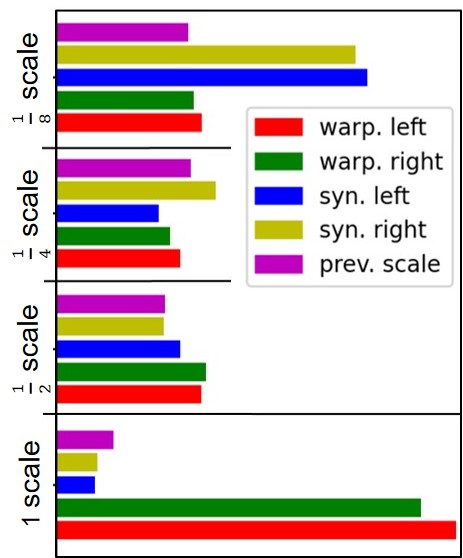}
    \caption{Average attention weights predicted by fusion network for synthesis and warping interpolation features on each scale.}
    \label{fig:average_weights}
\end{figure}

\emph{Multi-scale feature fusion}. We found that, by using shared synthesis encoders, we can get a 0.13 dB improvement, and, by applying multi-stage training, we can further boost performance by 1.06 dB in Tab.~\ref{tab:additional_ablations:fusion}. Additionally, in Fig.~\ref{fig:average_weights} we show the average attention weights predicted by our fusion network for synthesis and warping interpolation features on each scale. It is clear that the fusion network uses both synthesis and warping interpolation features, but prefers synthesis features on coarse scales and warping features on fine scales. This is consistent with observations in Time Lens~[27].

\begin{figure*}
    \includegraphics[width=\textwidth]{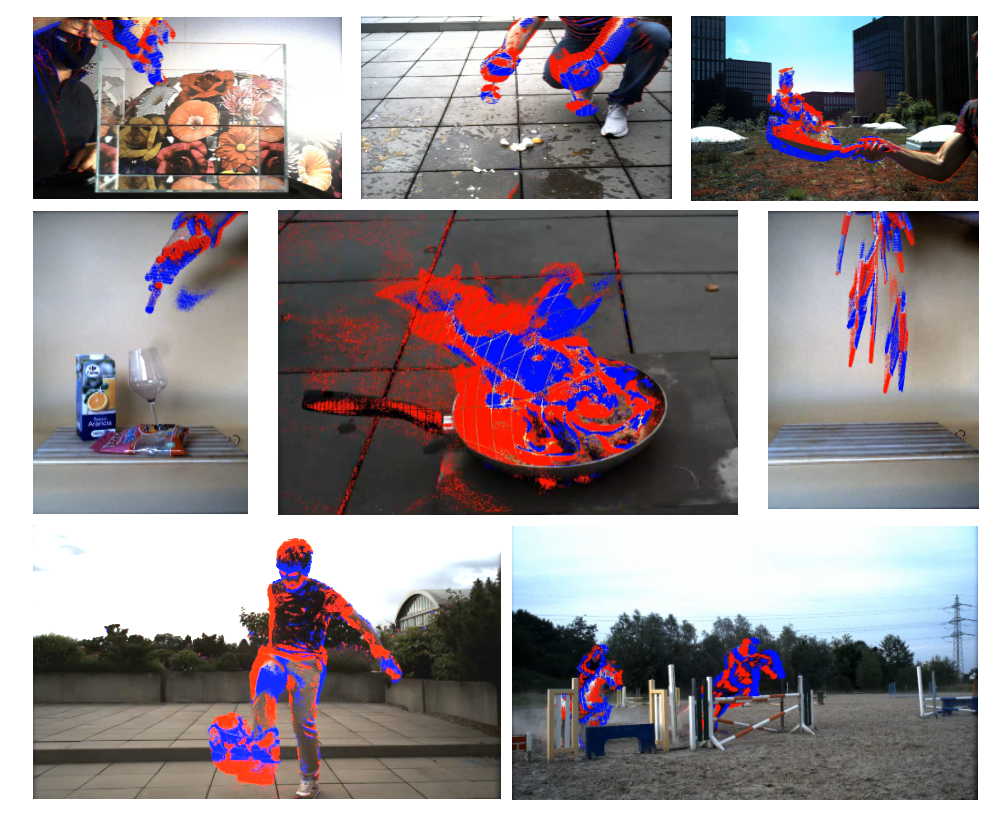}
    \caption[width=\textwidth]{The dataset includes challenging scene with water, color liquid, objects and liquid, small colorful objects, fire, thin colorful objects, rotating objects on a moving background, eye catching scene~(from top-left to bottom-right)}\label{fig:dataset_overview}%
\end{figure*}

\section{Beamsplitter Setup and Dataset}

\begin{figure}
    \centering
    \includegraphics[width=0.4\textwidth]{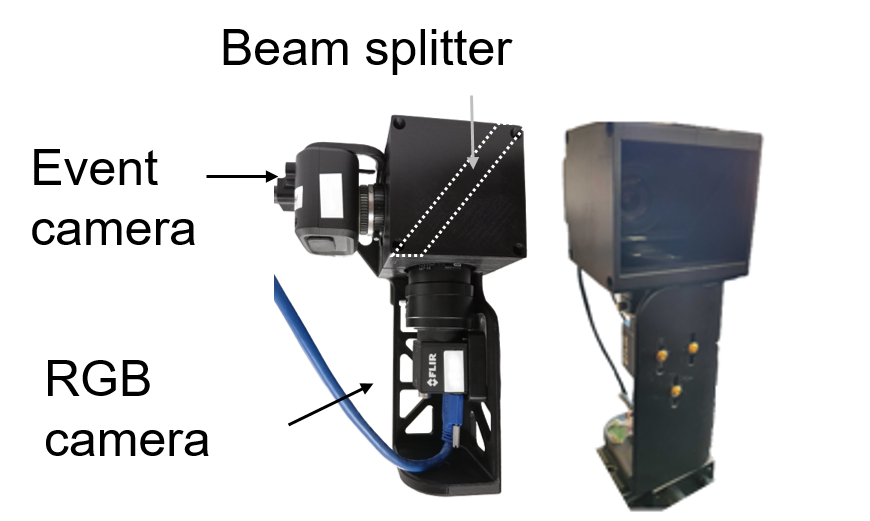}
    \caption{The beamsplitter setup mounts a FLIR RGB global shutter camera and a Prophesee Gen4 event camera\cite{Finateu20isscc} mounted on a case with a 50R/50T beam splitter mirror that allows the sensors to share a spatially aligned field of view. view}\label{fig:beamsplitter_setup}%
\end{figure}

We built an experimental setup with a global shutter RGB Flir 4096$\times$2196 camera and a Prophesee Gen4M 1280$\times$720 event camera\cite{Finateu20isscc} arranged with the beam splitter as shown in Fig.~\ref{fig:beamsplitter_setup}.

In our setup, the two cameras are hardware synchronized through the use of external triggers. 
Each time the standard camera starts and ends exposure, a trigger is sent to the event camera which records an \emph{external trigger event}
with precise timestamp information. This information allows us to assign accurate timestamps to the standard frames.

We use a standard stereo rectification procedure using images and event reconstructions from E2VID\cite{rebecq_2019b} to physically align the event and frame cameras and achieve a baseline of around 0.6mm. Next, we set the same focus for the event and frame camera and match the resolution of the RGB camera to the event camera by downsampling the images to a resolution of 1280$\times$720. Next, we calibrate each camera separately to estimate the lens distortion parameters and focal lengths. After removing the lens distortion, we estimate a homography that compensates for the misalignment between events and images. While this procedure removes misalignment for most of the scenes, small pixel misalignment still occurs for very close scenes due to the residual baseline. We automatically compensate for this misalignment for each sequence using a small global x-y shift, computed by maximizing the cross-correlation of event integrals and temporal image difference. Finally, after image acquisition, we adjust the brightness and contrast of the images. Using the procedure above, we collected 123 diverse and challenging scenes with varying depth, some of which we show in Fig.~\ref{fig:dataset_overview}. The dataset is divided into 78 training scenes, 19 validation scenes, and 26 test scenes.

{\small
\bibliographystyle{ieee_fullname}
\bibliography{main}
}

\end{document}